\documentclass{article} 

\def\iclrsubmission{0}

\if\iclrsubmission1
  \usepackage{iclr2024_conference,times}
\else
  \usepackage{preprint,times}
  \iclrfinalcopy 
\fi

\usepackage{amsmath,amsfonts,bm}

\def\eqref#1{equation~\ref{#1}}

\def\1{\bm{1}}

\DeclareMathAlphabet{\mathsfit}{\encodingdefault}{\sfdefault}{m}{sl}
\SetMathAlphabet{\mathsfit}{bold}{\encodingdefault}{\sfdefault}{bx}{n}

\DeclareMathOperator*{\argmax}{arg\,max}

\usepackage{hyperref}
\usepackage{url}
\usepackage{amsmath}
\usepackage{graphicx}
\usepackage{color}
\usepackage{xcolor}
\usepackage{siunitx}
\usepackage{amsfonts}
\usepackage{latexsym}
\usepackage{comment}
\usepackage{booktabs}
\usepackage{microtype}
\usepackage{xspace}
\usepackage{booktabs}
\usepackage{multirow}
\usepackage{dcolumn}
\usepackage{tikz}
\usepackage{bbding}
\usepackage{hhline}
\usepackage{mdframed}
\usepackage{enumitem}
\usepackage{xcolor}
\usepackage{hyperref}
\usepackage{url}
\newcolumntype{d}[1]{D{.}{.}{#1}}
\usepackage{makecell}
\usepackage{dirtytalk}
\usepackage{colortbl}
\usepackage{wrapfig}
\usepackage{caption}
\usepackage{bm}

\makeatletter
\newcommand\footnoteref[1]{\protected@xdef\@thefnmark{\ref{#1}}\@footnotemark}
\makeatother

\hypersetup{
  colorlinks   = true, 
  urlcolor     = blue!50!black, 
  linkcolor    = blue!50!black, 
  citecolor   = blue!50!black 
}

\title{ 
    \modelname\ Language Models: Isolating Legal Risk \\
    in a Nonparametric Datastore
}

\author{}

\definecolor{green}{rgb}{0.1,0.1,0.1}
\definecolor{chocolate}{HTML}{D2691E}
\definecolor{maroon}{HTML}{A00000}
\definecolor{indigo}{HTML}{4B0082}
\definecolor{green}{HTML}{008000}
\definecolor{red}{HTML}{a91e1e}
\definecolor{cadmiumgreen}{rgb}{0.0, 0.42, 0.24}
\definecolor{forestgreen}{rgb}{0.13, 0.55, 0.13}

\newcolumntype{L}[1]{>{\PreserveBackslash\raggedright}p{#1}}
\newcolumntype{R}[1]{>{\raggedleft\let\newline\\\arraybackslash\hspace{0pt}}m{#1}}
\newcolumntype{P}[1]{>{\centering\arraybackslash}p{#1}}

\usepackage{amssymb}
\usepackage{pifont}

\newcommand{\xmark}{\ding{55}}

\makeatletter
\newcommand*\myfontsize{%
  \@setfontsize\myfontsize{8}{9}%
}
\makeatother

\newcommand{\PD}{{\protect\color{chocolate} $\overline{\underline{\textsc{\texttt{pd}}}}$}}
\newcommand{\SW}{{\protect\color{blue!50} $\overline{\underline{\textsc{\texttt{sw}}}}$}}
\newcommand{\BY}{{\protect\color{indigo} $\overline{\underline{\textsc{\texttt{by}}}}$}}

\newcommand{\dataname}{\textsc{Open License Corpus}}
\newcommand{\datanameshort}{\textsc{OLC}}

\newcommand{\datanamePDSW}{\textsc{OLC} (\PD{}\SW{})}

\newcommand{\modelname}{\textsc{Silo}}

\newcommand{\tightparagraph}[1]{
    \vspace{-.5em} 
    \paragraph{#1}
}

\newcommand{\knnlm}{$k$NN-LM}
\newcommand{\knnlmbold}{$\bm{k}$NN-LM}
\newcommand{\riclm}{RIC-LM}

\usepackage{adjustbox}
\newcommand{\marktext}[2]{\adjustbox{bgcolor=#1}{\strut #2}}

\newcommand{\testPrefix}{\marktext{teal!30}{\textbf{\texttt{Test Prefix}}}}
\newcommand{\testContinuation}{\marktext{purple!30}{\textbf{\texttt{Test Continuation}}}}
\newcommand{\retrievedPrefix}{\marktext{green!30}{{\textbf{\texttt{Retrieved Prefix}}}}}
\newcommand{\retrievedContinuation}{\marktext{pink!70}{\textbf{\texttt{Retrieved Continuation}}}}
\newcommand{\cn}[1]{\textcolor{red}{\textbf{\underline{#1}}}}

\newcommand*\samethanks[1][\value{footnote}]{\footnotemark[#1]}

\newcommand{\myskip}[1]{}

\myskip{
\author{
Sewon Min\thanks{~Equal Contribution.} \\
  University of Washington \\
  \texttt{\small \href{mailto:sewon@cs.washington.edu}{\tt sewon@cs.washington.edu}} \\
  \and
  Suchin Gururangan\samethanks \\
  University of Washington \\
  \texttt{\small \href{mailto:sg01@cs.washington.edu}{\tt sg01@cs.washington.edu}} \\
  \and
  Eric Wallace\\
  UC Berkeley \\
  \texttt{\small \href{mailto:ericwallace@berkeley.edu}{\tt ericwallace@berkeley.edu}} \\
  \and
  Hannaneh Hajishirzi\\
  University of Washington \\
  \texttt{\small \href{mailto:hannaneh@cs.washington.edu}{\tt hannaneh@cs.washington.edu}} \\
  \and
  Noah A. Smith \\
  University of Washington \\
  \texttt{\small \href{mailto:nasmith@cs.washington.edu}{\tt nasmith@cs.washington.edu}} \\
  \and
  Luke Zettlemoyer \\
  University of Washington \\
  \texttt{\small \href{mailto:lsz@cs.washington.edu}{\tt lsz@cs.washington.edu}} \\
  }
}

\newcommand{\affilsup}[1]{\rlap{\textsuperscript{\normalfont#1}}}

\author{Sewon Min\affilsup{*1} \qquad
    Suchin Gururangan\affilsup{*1} \qquad
    Eric Wallace\affilsup{2} \qquad Weijia Shi\affilsup{1} \\
    \textbf{Hannaneh Hajishirzi}\affilsup{1,3} \qquad
    \textbf{Noah A. Smith}\affilsup{1,3} \qquad
    \textbf{Luke Zettlemoyer}\affilsup{1} \\
    $^1$University of Washington \quad
    $^2$UC Berkeley \quad $^3$Allen Institute for AI
    \\
    \texttt{\{sewon,sg01,swj0419,hannaneh,nasmith,lsz\}@cs.washington.edu} \quad
    \texttt{ericwallace@berkeley.edu}
}

\date{}

\footnotetext{Equal Contribution.
}

\ifx\useiclr\undefined
{}
\else
\iclrfinalcopy 
\fi

\newcommand{\nascomment}[1]{\textcolor{red}{[#1 -Noah]}}

\begin{document}

\maketitle

\begin{abstract}
The legality of training language models (LMs) on copyrighted or otherwise restricted data is under intense debate.
However,
as we show, model performance significantly degrades if trained only on low-risk text  (e.g., out-of-copyright books or government documents), due to its limited size and domain coverage.
We present \modelname{}, 
a new language model that manages this risk-performance tradeoff during inference. \modelname{} is built by
(1) training a parametric LM on the \dataname~(\datanameshort), a new corpus we curate with 228B tokens of public domain and permissively licensed text
(2) augmenting it with a more general and easily modifiable nonparametric datastore (e.g., containing copyrighted books or news) that is only queried during inference.
The datastore allows use of high-risk data without training on it,
supports sentence-level data attribution, and enables data producers to opt out from the model by removing content from the store. These capabilities can foster compliance with data-use regulations such as the \emph{fair use} doctrine in the United States and the GDPR in the European Union.
Our experiments show that the parametric LM struggles on domains not covered by \datanameshort. However, access to the datastore greatly improves out of domain performance,
closing 90\% of the performance gap with an LM trained on the Pile, a more diverse corpus with mostly high-risk text.
We also analyze which nonparametric approach works best, where the remaining errors lie, and how performance scales with datastore size. Our results suggest that it is possible to build high quality language models while mitigating their legal risk.\footnote{We release all models, data, and code publicly at \url{https://github.com/kernelmachine/silo-lm}.} 
\end{abstract}

\section{Introduction}\label{sec:intro}Large language models (LMs) are under widespread legal scrutiny, in large part because they are trained on copyrighted content, which may infringe on the rights of data producers~\citep{metz2022lawsuit,vincent2023,jlversusalphabet,brittain2023copyright}. 
At the heart of this discussion is the inherent tradeoff between legal risk and model performance.
Training only on data sources such as public domain, non-copyrightable or otherwise permissively licensed data significantly degrades performance (as we show in \S\ref{sec:data}).
This limitation arises from the  scarcity of permissive data and its narrow specificity to sources such as copyright-expired books, government documents, and permissively licensed code, which are largely different from common LM corpora that cover more diverse domains~\citep{raffel2020exploring,gao2020pile,together2023redpajama}.

\begin{figure*}[t]
\centering
\includegraphics[trim={3.5cm, 3.5cm, 0.25cm, 1cm}, clip,width=0.93\linewidth]{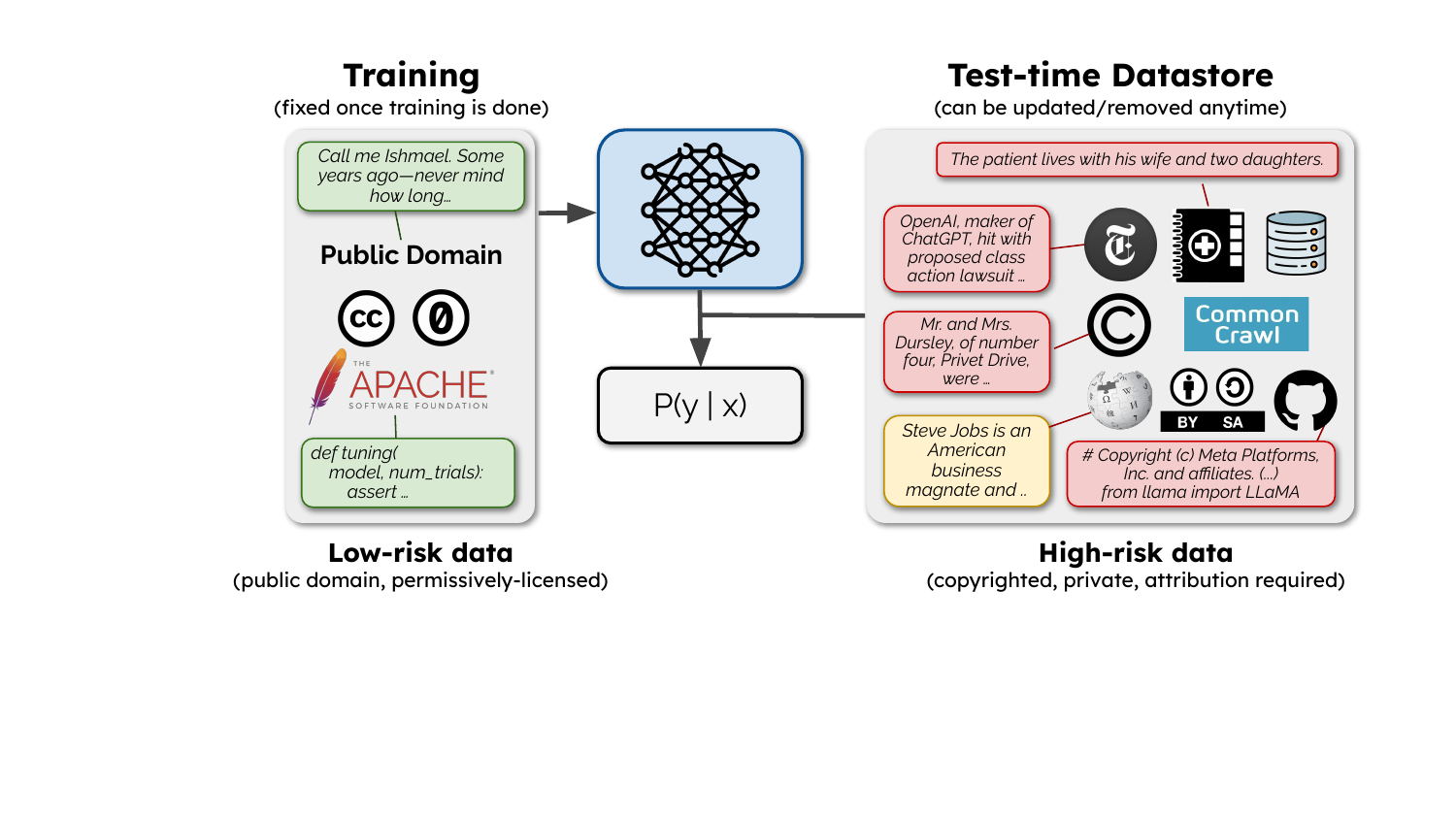}\vspace{-.5em}
\caption{\textbf{An overview of \modelname{}.} We train a {\em parametric} language model on low-risk datasets that contain public domain text (e.g., copyright-expired books) and permissively licensed code.
At inference time, we use a \emph{nonparametric datastore} that can include high-risk data, including medical text with personally-identifiable information,
copyrighted news,
copyrighted books, data requiring attribution, and code under non-permissive licenses (counterclockwise from the top of figure). The datastore can be modified at any time, e.g., to respond to opt-out requests. }
\label{fig:overview}
\end{figure*}

In this paper, we demonstrate it is possible to
improve the risk-performance tradeoff
by segregating training data into two distinct parts of the model: parametric and nonparametric  (Figure~\ref{fig:overview}). We learn LM parameters on \emph{low-risk} data (i.e., data under the most permissive licenses), and then use \emph{high-risk} data (i.e., data under copyright, restrictive licenses, or unknown licenses) in an inference-time-only
nonparametric component (called a {\em datastore}). 
With nonparametric datastores, we can \emph{retrieve} high-risk data to improve model predictions without training on it. The datastore can be easily updated at any time, and allows creators to remove their data from the model entirely, at the level of individual examples. 
This approach also attributes model predictions at the sentence-level, enabling credit assignment to data owners. 
These new capabilities enable better alignment of the model with various data-use regulations, e.g., the \emph{fair use} doctrine in the United States~\citep{henderson2023foundation} and the GDPR in the European Union~\citep[][]{zhang2023right}, as detailed in \S\ref{sec:background}.
This is in contrast to parametric models, where removing high-risk data is infeasible after training~\citep{bourtoule2020machine,carlini2021extracting} and data attribution at scale is difficult~\citep{
zhang2021counterfactual,han2023understanding}.

\myskip{
Unlike parametric models where removing high-risk data is infeasible after training~\citep{bourtoule2020machine,carlini2021extracting},
a nonparametric datastore offers flexibility as it can be easily updated and added, and allows for data opt-out at the level of individual examples. 
Furthermore, this approach attributes the model's predictions for free, which can facilitate proper credit assignment to data owners. \nascomment{feels like you should mention key papers introducing the nonparametric LM idea here, even if they're a bit older.  right now readers might think you think you're inventing that}
}

We introduce \textbf{\modelname},
a new nonparametric language model that 
follows our proposal
(\S\ref{sec:model}).
The parametric component in \modelname\ is trained on a new pretraining corpus, the \textbf{\dataname} (\textbf{\datanameshort}, \S\ref{sec:data}), which we curate to include data
under three types of permissive licenses, from public domain to Creative Commons.
\datanameshort\ is diverse but has a domain distribution that is very different from typical pre-training corpora; it is dominated by code and government text.
This leads to a new 
challenge of generalizing a model trained on highly specific domains, which we call {\em extreme domain generalization}.
We train three 1.3B-parameter LMs on varying 
subsets of \datanameshort, and 
then construct a test-time datastore that can include high-risk 
data, employing a retrieval method to make use of the datastore's contents during inference. 
We compare two widely studied retrieval methods: a nearest-neighbors approach (\knnlm) that uses a nonparametric next-token prediction function~\citep{khandelwal2020generalization} and a retrieval-in-context approach (\riclm) that retrieves text blocks and feeds them to the parametric LM in context \citep{shi2023replug,ram2023context}.

We evaluate \modelname{} in language modeling perplexity on 14 different domains, covering both in-domain and out-of-domain data with respect to \datanameshort~(\S\ref{sec:exp}).
These domains highlight specific legal risks,
e.g., copyrighted materials such as books, news and user reviews, or private data such as emails and clinical notes.
We compare \modelname{} to Pythia~\citep{biderman2023pythia}, a parametric LM with a similar parameter count but trained mostly on high-risk data~\citep{gao2020pile}.\footnote{The Pile contains a large amount of copyrighted or restrictively licensed data,
e.g., most content in its Books3, ArXiv, Github, OpenWebText, YoutubeSubtitles, and Common Crawl subsets.}
We first show that parametric-only \modelname~is competitive on domains covered by \datanameshort\ but falls short out-of-domain, confirming the challenge of extreme domain generalization. 
However, adding an inference-time 
datastore to \modelname~effectively addresses this challenge. 
Comparing the two methods of retrieving over this datastore, we find that while both \knnlm\ and \riclm\ significantly improve out-of-domain performance,
the former generalizes better than the latter,
allowing \modelname{} to reduce the gap with the Pythia baseline  by 90\% on average across all domains. 
%
Further analysis attributes these improvements to two factors: 
(1) \knnlm\ strongly benefits from scaling the datastore and (2) the nonparametric next-token prediction in \knnlm\ is robust to domain shift.
Altogether, our study suggests 
that in the few domains where \modelname{} has not yet matched Pythia performance levels, the remaining gaps can likely be closed by scaling the datastore size and further enhancing the nonparametric model.
\section{Background \& Related Work}\label{sec:background}\paragraph{Training datasets for language models.} State-of-the-art LMs are trained on vast text corpora that consist of billions or even trillions of tokens~\citep{gpt3,raffel2020exploring,gao2020pile,together2023redpajama}. These training sets are built by combining (1) manually selected sources such as Wikipedia, book collections, and GitHub and (2) web pages collected through web-crawling services such as Common Crawl. Most LM training efforts ignore copyright and intellectual property regulations that apply to these texts. For example, sources such as GitHub repositories and book collections typically contain text with highly restrictive licenses~\citep{bandy2021addressing}.

\tightparagraph{Legality of language models.}
The legality of training LMs this way has become a subject of intense debate, 
with numerous lawsuits being filed 
in the United States, United Kingdom, and beyond~\citep{gershgorn2021,metz2022lawsuit,vincent2023,de2023chatgpt,silvermanmeta,jlversusalphabet,silverman,tremblay}.  While the outcome of the lawsuits is uncertain, it is likely that such legal issues will continue to be a major factor in future LMs, especially since each country has its own data regulations. For example,


\begin{itemize}[itemsep=2pt, leftmargin=8pt, topsep=5pt]
    \item In the United States, the \emph{fair use doctrine} allows the public to use copyrighted data in certain cases, even without a license \citep{henderson2023foundation}. Deciding whether or not a model's use of copyrighted work constitutes fair use involves multiple dimensions, including whether the trained model is intended for commercial use, whether or not the work is factual or creative, the amount of the copyright content used, and the value of the copyrighted work. 
    There are claims that using parametric language models for \emph{generative} use-cases does \emph{not} constitute fair use, because the technology may output the copyrighted text verbatim~\citep{lemley2020fair}, which also has been shown empirically~\citep{carlini2021extracting,carlini2022quantifying,kandpal2022deduplicating,chang2023speak}.
    This is in contrast to \emph{transformative} technologies, such as classifiers, which may use the copyrighted text but do not directly generate content, which the fair use doctrine favors.
    We refer readers to \citet{henderson2023foundation} 
    for a more comprehensive discussion. 
    \item The {\em General Data Protection Regulation (GDPR)} is a comprehensive data protection and privacy law in the European Union (EU). It grants individuals more control over their data by regulating organizations and businesses.
    The obligations include (1) obtaining consent from users before processing their data, (2) providing transparency about data processing, (3) ensuring data security, and (4) allowing individuals to access, correct, and erase their data.
    GDPR has global impact, as many international companies handle EU citizens' data.
    While it is under debate how GDPR is applied to training language models, compliance with GDPR is  expensive (e.g., requiring retraining for every data correction or erasure).
    See \citet{zhang2023right} for more discussion on challenges for compliance with
    the GDPR's {\em Right to Erasure}  (and the {\em Right to be Forgotten} in general).
\end{itemize}
The goal of our work is not to weigh in on legal discussions; instead, we study the feasibility of developing technologies that explicitly manage legal risk.
In particular, our technique places all copyrighted data in a nonparametric datastore.
While the data is still used in service of a generative model, restricting copyrighted data in a datastore and providing instance-level attribution and data opt-out can increase the likelihood of a successful fair use defense~\citep{henderson2022pile}.\footnote{
    Our model on its own does not entirely remove legal risk. Rather, it provides functionalities that, when used appropriately, lower legal risk and strengthen a fair use defense. See \S\ref{sec:discuss} for a discussion.
} Moreover, GDPR's requirement regarding user data access, correction, and erasure aligns well with the capabilities of the datastore.
Attribution and opt-out are fundamental features of our model (\S\ref{subsec:non-parametric-lm}). This is in contrast to other techniques like post-hoc training data attribution~\citep{koh2017understanding,han2023understanding} and the removal of the effect of particular training examples from parameters~\citep{cao2015towards,jang2023knowledge}, which lack inherent guarantees and are hard to scale.


\tightparagraph{Prior work in copyright risk mitigation.}
The most straightforward approach to avoid copyright infringement is to filter training data to only include permissive licenses.
This has been done in prior work, primarily for code-based datasets \citep[e.g.,][]{kocetkov2023the, fried2023incoder, together2023redpajama} and scientific text \citep[e.g.,][]{peS2o}. Extending a similar approach to a wider range of domains remains unclear, because  permissive data is extremely scarce in most domains, e.g., books and news.
For the same reason, \citet{henderson2023foundation} has suggested that restricting the training data to public domain or otherwise permissively licensed data may be impractical. In this work, we show that there is in fact a large number of tokens from data sources with permissive licenses, but the key challenge instead arises from the highly skewed domain distribution. See \S\ref{sec:discuss} for other copyright mitigation strategies that are more technical in nature.

\section{Building the \dataname: A Permissively-Licensed Pre-training Corpus}\label{sec:data}

Our study focuses on addressing the legal risk of copyright violation in language models by separating \emph{low-risk} data sources (i.e., those in the public domain or under permissive licenses) from \emph{high-risk} ones (i.e., those with unknown licenses or under copyright).
We introduce the \textbf{\dataname\ (\datanameshort)}, a large collection of permissive textual datasets across multiple domains with a taxonomy of data licenses that delineate their permissiveness (\S\ref{subsec:data-definition}). We group the data into three levels of legal permissiveness (\S\ref{subsec:our-data}) and conduct a thorough analysis (\S\ref{subsec:our-data-analysis}).
This curated data is then used to train model parameters (\S\ref{sec:model}) and highlights the challenge of extreme domain generalization due to its skewed domain distribution.

\tightparagraph{A disclaimer.}
The license taxonomy and categorization of texts that we present
is by no means perfect, and \datanameshort~should not be considered a universally safe-to-use dataset.  
The license
associated with a document may be time- and country-dependent, e.g., Gutenberg books~\citep{gutenberg} are public domain in the United States, but some of them may still have copyright attached outside of the United States.  Moreover, other legal constraints (e.g., the Digital Millenium Copyright Act)\footnote{\url{https://www.copyright.gov/dmca/}} may prohibit the use of a data source despite a permissive data license. Finally, we do not explicitly filter out personally identifiable information from the corpus, so it is possible that certain subsets still pose privacy risks despite being permissively licensed. We encourage users of \datanameshort~to consult a legal professional on the suitability of each data source for their application.

\subsection{Taxonomy of Data Licenses}\label{subsec:data-definition}

As discussed in \S\ref{sec:background}, determining what data one is permitted to use from a copyright perspective is an ongoing topic of debate, and is context- and country-dependent~\citep{henderson2023foundation}. In this paper, we take a conservative approach where we train models using only text with the most permissible licenses, thus enabling widespread downstream use. 
Concretely, we focus on 
four broad categories:

\begin{itemize}[itemsep=2pt, leftmargin=8pt, topsep=1pt]
\item \textbf{Public domain} (\PD) text has no restrictions. This includes texts whose intellectual property rights have expired (e.g., the works of William Shakespeare) or been expressly waived by the creator (e.g., CC0-licensed scientific papers). 
\item \textbf{Permissively licensed software} (\SW{}) including MIT, Apache, and BSD software are quite permissive to use. Unlike public domain text, these licenses typically carry some basic stipulations such as requiring one to include a copy of the original license (although, it is debatable whether it is still required when the associated text is used as data or treated as a software).
The code is otherwise free to use, and code is generally well protected by fair use clauses~\citep{lemley2020fair}. 
\item \textbf{Attribution licenses} (\BY{}) such as Creative Commons Attribution (CC-BY) 
are free to use as long as \say{credit is given to the creator.}
For example, if a journalist quotes an article from Wikipedia (a CC-BY source), then they must provide a form of citation, link, or attribution back to the original source. In the context of machine learning, it is not clear what an attribution would constitute. For example, under one interpretation, every LM generation should include a complete list of sources that contributed highly to it~\citep{henderson2023foundation}. In this paper, we take a conservative approach and do not include \BY\ data in the main experiments, but still include the \BY\ data for future use as well as for ablations, since \BY\ data is generally considered quite permissive. 
\item \textbf{All other data} that is not in one of the above three categories is assumed to be non-permissive. This includes: any text that is explicitly protected by copyright or licenses that are non-commercial (e.g., CC-NC), any software without clear MIT, BSD, or Apache licenses, and any generic web-crawled data where the license or copyright information may be unclear. 
\end{itemize}

In \S\ref{subsec:data-allocation}, we train the models on varying subsets of licenses---from \PD\ and \PD\SW\ to \PD\BY\SW---to accommodate different risk tolerances.



\begin{table}[t]
    \myfontsize 
    \setlength\extrarowheight{-3pt}
    \centering
    \begin{tabular}{lllr}
        \toprule
            \textbf{Domain} & \textbf{Sources} & \textbf{Specific License} & \textbf{\# BPE Tokens (B)} \\
        \cmidrule(lr){1-4}
            \multirow{2}{*}{Legal} & \PD\ Case Law, Pile of Law (PD subset) & Public Domain & 27.1\hphantom{0} \\[0.5ex]
             & \BY\ Pile of Law (CC BY-SA subset) & CC BY-SA & 0.07 \\
        \cmidrule(lr){1-4}
            Code & \SW\ Github (permissive) & MIT/BSD/Apache & 58.9\hphantom{0} \\
        \cmidrule(lr){1-4}
            \multirow{2}{*}{Conversational} & \SW\ HackerNews, Ubuntu IRC & MIT/Apache & 5.9\hphantom{0} \\[0.5ex]
            & \BY\ Stack Overflow, Stack Exchange & CC BY-SA & 21.3\hphantom{0} \\
        \cmidrule(lr){1-4}
            Math & \SW\ Deepmind Math, AMPS & Apache & 3.5\hphantom{0}  \\
        \cmidrule(lr){1-4}
            \multirow{2}{*}{Science} & \PD\ ArXiv abstracts, S2ORC (PD subset) &  Public Domain &1.2\hphantom{0} \\[0.5ex]
            & \BY\ S2ORC (CC BY-SA subset) & CC BY-SA & 70.3\hphantom{0} \\
        \cmidrule(lr){1-4}
            Books & \PD\ Gutenberg  & Public Domain & 2.9\hphantom{0} \\
        \cmidrule(lr){1-4}
            \multirow{2}{*}{News} & \PD\ Public domain news & Public Domain & 0.2\hphantom{0} \\[0.5ex]
            & \BY\ Wikinews & CC BY-SA & 0.01\\
        \cmidrule(lr){1-4}
            Encyclopedic & \BY\ Wikipedia & CC BY-SA & 37.0\hphantom{0} \\
        \bottomrule
    \end{tabular}
    \vspace{-0.1cm}
    \caption{\textbf{Overview statistics of \datanameshort{}}. 
    \PD, \SW, and \BY\ indicates public domain data, data under permissive software licenses, and data under attribution licenses, respectively.
    Some corpora contain a mixture of different licenses (e.g., Pile of Law and S2ORC), which we split into subsets based on 
    per-document licenses.
    BPE tokens are based on the GPT-NeoX tokenizer \citep{black2022gptneox20b}.  
    }\label{tab:data}
\end{table}

\subsection{Building the \dataname}\label{subsec:our-data}

Based on this taxonomy of licenses, OLC is a 228B token corpus of \PD{}, \SW{}, and \BY{} data. \datanameshort~consists of 17 manually-selected sources of 
primarily English text that are under permissive licenses,\footnote{We include the data in only when the license information is clearly stated as part of metadata. While we tried our best to collect the data for \datanameshort, it is possible we missed potential sources, as it relies on manual efforts; future work can study collecting more permissive text at scale, as discussed in \S\ref{sec:discuss}.} as summarized in Table~\ref{tab:data}.

The text generally falls into eight different domains:

\begin{itemize}[itemsep=2pt, leftmargin=8pt, topsep=1pt]
\item \PD\ \BY\ \textbf{Legal:} We curate legal text from the Pile of Law~\citep{henderson2022pile}, an amalgation of 31 different sources of text related to civil court cases, patents, and other legal and governmental works, either licensed as public domain or CC-BY.
We also gather public domain text from the Case Law Access Project~\citep{caselaw2018}, which covers over 6.5 million decisions published by state and federal courts throughout U.S. history. 

\item \SW\ \textbf{Code:} We use the Github subset of the RedPajama dataset \citep{together2023redpajama}, which contains code from Github repositories with three permissive software licenses: MIT, Apache, and BSD. 

\item \SW\ \BY\ \textbf{Conversation:} We source conversational text under permissive software licenses from the HackerNews (MIT license) and the Ubuntu IRC (Apache license) subsets of the Pile~\citep{gao2020pile}. We also use the Stackexchange subset of the RedPajama dataset \citep{together2023redpajama} and a Stackoverflow corpus from Kaggle,\footnote{\url{https://www.kaggle.com/datasets/stackoverflow/stackoverflow}} both under the CC-BY-SA license.

\item \SW\ \textbf{Math:} We source mathematical text from the Deepmind Mathematics~\citep{saxton2019analysing} and the AMPS~\citep{hendrycksmath2021} datasets, both of which are under the Apache license.

\item \PD\ \BY\ \textbf{Science:} We source scientific text from ArXiv abstracts that are in the public domain~\citep{arxiv_dataset}. We also collect full-text articles from the Semantic Scholar Research Corpus \citep[S2ORC]{lo-etal-2020-s2orc}, either licensed as public domain or CC-BY.

\item \PD\ \textbf{Books:} We source books from the Gutenberg corpus~\citep{gutenberg}, which are copyright-expired books that are in the public domain.

\item \PD\ \BY\ \textbf{News:} We collect public domain news text from the English subset of the MOT corpus~\citep{palenmichel2022multilingual}. We also collect text from Wikinews, which is under CC BY-SA.

\item  \BY\ \textbf{Encyclopedic:} Finally, we include a large set of Wikipedia from the subset included in RedPajama~\citep{together2023redpajama}.
We follow RedPajama in using Wikipedia snapshots from 20 languages even though the model primarily focuses on English.
\end{itemize}

\begin{table}[t]
    \centering \myfontsize
    \setlength{\tabcolsep}{5pt}
    \begin{tabular}{lrrrrrrrr}
        \toprule
            & \multicolumn{2}{c}{\PD{}}  & \multicolumn{2}{c}{\PD{}\SW{}} & \multicolumn{2}{c}{\PD{}\SW{}\BY{}} & \multicolumn{2}{c}{\emph{The Pile}} \\
            \cmidrule(lr){2-3} \cmidrule(lr){4-5} \cmidrule(lr){6-7} \cmidrule(lr){8-9}
            \textbf{Domain} & \textbf{Tokens (B)} & \textbf{\%} & \textbf{Tokens (B)} & \textbf{\%} & \textbf{Tokens (B)} & \textbf{\%} & \textbf{Tokens (B)} & \textbf{\%} \\
        \midrule
            Code            &0.0&0.0& 58.9  & 59.1 & 58.9 & 25.8 & 32.6 & 9.8 \\
            Legal           &27.1&86.2 & 27.1  & 27.2 & 27.2 & 11.9 & 30.8 & 9.3  \\
            Conversation    &0.0&0.0& 5.9   & 5.9 & 27.2 & 11.9 & 33.1	& 10.0\\
            Math            &0.0&0.0& 3.5   & 3.5 & 3.5 & 1.50 & 7.1 & 2.1 \\
            Books           &2.9&9.3& 2.9   & 2.9 & 2.9 & 1.3 & 47.1 & 14.2 \\
            Science         &1.2&3.8& 1.2   & 1.2 & 71.5 & 31.3 & 86.0 &	26.0 \\
            News            &0.2&0.7& 0.2 & 0.2 & 0.2 & 0.1 & -$^\dagger$ & -$^\dagger$ \\
            Wikipedia       &0.0&0.0& 0.0   & 0.0 & 37.0 & 16.2 & 12.1 &	3.7 \\
            Unverified web  &0.0&0.0& 0.0 & 0.0 & 0.0 & 0.0  & 83.1	& 25.0 \\
        \midrule
            Total           &31.4&100.0& 99.6 & 100.0 & 228.3 & 100.0 & 331.9 & 100.0 \\
        \bottomrule
    \end{tabular}
    \caption{\textbf{\datanameshort\ is large but its distribution is different from that of typical pretraining corpora like the Pile.} Data distribution of \datanameshort\ (\PD, \PD\SW, \PD\SW\BY) in comparison to the  
    Pile~\citep{gao2020pile}, a common LM training dataset that is not specifically designed for legal permissibility. We report the number of tokens in billions, and the relative frequency.
    $\dagger$: There is no explicit news domain in the Pile, but news sites are found to be some of the most representative data sources in Common Crawl~\citep{dodge2021documenting}.
    }\label{tab:data-distribution}
\end{table}


\vspace{.8em}
Following \citet{kandpal2022deduplicating,lee2021deduplicating}, we deduplicate text using \citet{bff}, a document-level filter that considers $n$-gram overlap.
We first deduplicate within each domain to remove redundant documents from similar sources (e.g. Case Law and the Pile of Law), and then perform deduplication against the validation and test datasets of the Pile to avoid test leakage.

\subsection{Analysis of \datanameshort}\label{subsec:our-data-analysis}

In Table \ref{tab:data-distribution}, we compare the distribution of domains in \datanameshort~to that of the Pile \citep{gao2020pile}, a popular pretraining corpus that includes data under copyright restrictions (e.g., Books, web crawl).\footnote{This comparison also dovetails with our experiments in \S\ref{sec:exp}, where we compare \modelname~to Pythia, a model trained on the Pile.}  These statistics convey a number of research challenges when working with \datanameshort.
First, while we tried our best to collect public domain or permissively-licensed data, the size of \datanameshort\ is still 31\% smaller than the Pile.
In addition, while the majority of the Pile is sourced from scientific text, web crawl, and books, \datanameshort~is dominated by code, scientific text, and legal text. This highlights that models designed for use outside these specific domains will likely struggle and may require special techniques for extreme domain generalization.

To analyze this further, we perform an $n$-gram based analysis of \datanameshort~domains against the validation data of the Pile, to better understand the domain shifts. For each validation domain, we examine the \emph{maximum} $n$-gram overlap across all \datanameshort~domains (see \S\ref{app:additional-data-analysis} for more details).
\datanameshort~domains have substantially less overlap with the validation data as compared to the Pile training domains: on average, the overlap between \datanameshort~domains and the validation domains is just 17\%$\pm$9\%, versus 28\%$\pm$14\% for the Pile training data.
However, we find a large variance in overlap statistics across domains in \datanameshort; we display the full matrix of $n$-gram overlap in \S\ref{app:additional-data-analysis}. These results provide further evidence that models trained on \datanameshort\ must handle larger domain shifts at test time than models trained on the Pile. Later, we show that these $n$-gram overlap statistics correlate strongly with language modeling performance (\S\ref{subsec:results-parametric}).

\section{\modelname{} 
}\label{sec:model}\begin{figure*}[t]
\centering
\includegraphics[trim={2.9cm, 4.5cm, 3.8cm, 2cm}, clip,width=1.0\linewidth]{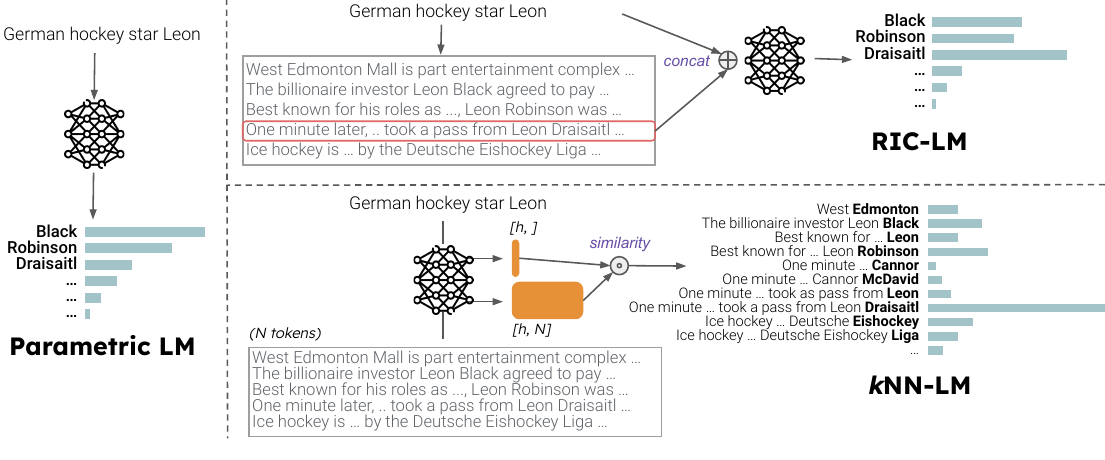}\vspace{-.5em}
\caption{
    \textbf{An illustration of a parametric model and two retrieval methods we compare: \riclm\ and \knnlmbold.}
    The orange boxes indicate representations of the input prefix and the tokens in the datastore, each in $\mathbb{R}^h$ and $\mathbb{R}^{h \times N}$, where $h$ is a hidden dimension and $N$ is the number of tokens in the datastore.
    The distribution from \knnlm\ in the figure describes $P_{k\mathrm{NN}}$;
    while omitted in the figure, the final output distribution from \knnlm\ is an interpolation between $P_{k\mathrm{NN}}$ and the distribution from the parametric LM.
    See \S\ref{subsec:non-parametric-lm} for more details of each method.
}
\label{fig:method}
\end{figure*}

We introduce \modelname{}, which combines an LM trained on permissive data with a nonparametric datastore based on less restricted data. Our goal with \modelname{} is to build an LM---i.e., a model that takes a prefix of text
$x$ and outputs a next-word probability distribution over the vocabulary $P(y \mid x)$---but to do so in a legally safe way.
%
We first describe the general methodology from prior work (\S\ref{subsec:parametric-lm}--\ref{subsec:non-parametric-lm}) and then how we build \modelname{} upon them by placing low-risk data and high-risk data to model parameters and a nonparametric datastore, respectively (\S\ref{subsec:data-allocation}).
Implementation details are provided in \S\ref{subsec:exp-setup}.

\subsection{The Parametric Component}\label{subsec:parametric-lm}

For the parametric component of \modelname, we use a standard, dense, decoder-only transformer LM~\citep{vaswani2017attention} using the LLaMA architecture~\citep{touvron2023llama}.
This model uses a fixed set of parameters at both training and inference time.


\subsection{The Nonparametric Component}\label{subsec:non-parametric-lm}


We experiment with two widely-used retrieval methods for the nonparametric component (Figure \ref{fig:method}): the $k$-nearest neighbors LM (\knnlm; \citealp{khandelwal2020generalization}) and the retrieval-in-context approach (\riclm; \citealp{shi2023replug,ram2023context}). 
Each approach constructs a datastore from the raw text data offline, and then uses it on-the-fly at inference time.


\tightparagraph{The $\bm{k}$-nearest neighbors language model (\knnlmbold).} A \knnlm~\citep{khandelwal2020generalization} interpolates the next-token probability distribution from a parametric LM with a nonparametric distribution based on every token that is stored in a datastore.
Given a text dataset consisting of $N$ tokens $c_1...c_N$, a datastore is built by creating 
a key-value pair for every token $c_i$ ($1 \leq i \leq N$). Specifically, a value is $c_i$ and a key $k_i$ is $...c_{i-1}$, a prefix preceding $c_i$.
%
%
At test time, given an input prefix $x$, the nonparametric distribution is computed by:
$$
    P_{k\mathrm{NN}}(y\mid x) \propto \sum_{(k,v) \in \mathcal{D}} \mathbb{I}[v=y] \left(-d(\mathrm{Enc}(k), \mathrm{Enc}(x))\right).
$$
Here, $\mathrm{Enc}$ is an encoder that maps a text into $\mathbb{R}^h$ and $d: \mathbb{R}^h \times \mathbb{R}^h \rightarrow \mathbb{R}$ is a distance function, where $h$ is the hidden dimension. We follow \citet{khandelwal2020generalization} and use the output vector from the last layer of the transformers in the parametric LM as $\mathrm{Enc}$, L2 distance as $d$, and an approximate nearest neighbor search using FAISS~\citep[details in \S\ref{subsec:exp-setup}]{johnson2019billion}.
The final model takes the \knnlm\ output and interpolates it with the output from the parametric LM:\footnote{
    While the encoder that outputs $P_\mathrm{k\mathrm{NN}}(y\mid x)$ and the parametric LM that outputs $P_\mathrm{LM}(y\mid x)$ are based on the same transformer models in this case following \citet{khandelwal2020generalization}, it is not a necessary condition. One of our ablations in \S\ref{subsec:results-nonparametric} use different transformer models for the encoder 
    and the parametric LM.
} $\lambda P_\mathrm{LM}(y\mid x) + (1-\lambda) P_{k\mathrm{NN}}(y\mid x),
$ where $\lambda$ is a fixed hyperparameter between $0$ and $1$.

Future work can improve \knnlm, e.g.,
by training the model to output a nonparametric distribution~\citep{zhong2022training,lan2023copy,min2022nonparametric}, by having a vocabulary-specific $\lambda$~\citep{huang2023k}, 
or by modeling $\lambda$ as a function of the input $x$~\citep{he-etal-2021-efficient,drozdov2022you}.


\tightparagraph{The retrieval-in-context language model (\riclm).}


As an alternative to $k$NN-LM,  \riclm\ \citep{shi2023replug,ram2023context} retrieves text blocks from a datastore and feeds them to the parametric LM in context. Specifically, given a dataset consisting of $N$ tokens $c_1...c_N$, an index $\mathcal{D}$ is constructed by splitting the data into text blocks $b_1...b_M$, optionally with a sliding window.
%
%
At test time, given an input prefix $x$, \riclm\ retrieves the most similar paragraph to the prefix $\hat{p} = \argmax_{b \in \mathcal{D}} \mathrm{sim}(b, x)$ and concatenates it to the prefix to produce $P_\mathrm{LM}(y\mid \hat{b}, x)$.
Here, $\mathrm{sim}$ is a function that computes a similarity score between two pieces of text; we use BM25 following \citet{ram2023context} who show that BM25 outperforms alternative dense retrieval methods. 

Future work can improve \riclm, e.g., by using multiple text blocks through ensembling~\citep{shi2023replug} or reranking~\citep{ram2023context}, by tuning the retrieval system~\citep{shi2023replug}, or by training the LM to use retrieved blocks in context~\citep{guu2020retrieval,izacard2022few}.

\tightparagraph{Comparison between \knnlmbold~and \riclm.}
The key difference between \knnlm\ and \riclm\ lies in how the nonparametric component influences the output. In \knnlm, it directly impacts the output distribution, while in \riclm, it indirectly influences the output by affecting the input to the parametric model.
\knnlm\ intuitively benefits more from a datastore as it provides direct influence to the output and relies less on the parametric component.
Nonetheless, \riclm\ interacts more easily with a parametric model (i.e., it is applicable to a black-box LM) and offers better speed and memory efficiency (explored in Appendix~\ref{app:additional-nonparametric}).

Empirical comparisons between kNN-LM and RIC-LM have been largely unexplored; in fact, we are unaware of such work.
In our experiments (\S\ref{subsec:results-nonparametric}), we present a series of such comparisons, with varying sizes of the datastore, and with and without distribution shift.

\tightparagraph{Attribution and opt-out.}
Since elements in the datastore that contribute to the model prediction are transparent, both \knnlm\ and \riclm\ offer inherent attributions. Moreover, 
data removed from the datastore is guaranteed not to contribute to any model predictions, allowing data owners to remove their data at the level of individual examples.
Both are unique characteristics of nonparametric language models.
While prior work studies post-hoc attribution to the data used for training model parameters~\citep{koh2017understanding,han2023understanding} and removing the effect of specific training examples from parameteric models~\citep{cao2015towards,jang2023knowledge}, they
are arguably not fundamental due to lack of inherent guarantees, and are difficult to scale.



\subsection{Building \modelname}\label{subsec:data-allocation}
\modelname\ is is built upon the general methodology of \knnlm\ and \riclm. However, unlike prior work that uses the same data for learning model parameters and a nonparametric datastore, \modelname\ uses distinct datasets for these two components.

The key idea behind \modelname\ is to use low-risk data to estimate model parameters, and to use high-risk data only in a nonparametric datastore. 
This is based on the motivation that model parameters should be learned conservatively, since training data is difficult to remove or trace after model training is completed.
In contrast, a nonparametric datastore offers greater flexibility, as it can be easily updated, grown, or filtered, supports data opt-out at the level of individual examples, and provides attributions for free to every model prediction. 
These functions enable adherence to data-use regulations (\S\ref{sec:background}). 


\tightparagraph{Training datasets.} We train each of our LMs on one of the three datasets of \datanameshort: \PD{} data, \PD{}\SW{} data, and \PD{}\SW{}\BY{} data. Each of the resulting models constitutes a different level of possible copyright infringement risk. 

\tightparagraph{Datastore.}
We assume in-distribution data for each test domain is available at inference time, and construct a datastore for each domain (details in \S\ref{subsec:exp-setup}). Future work may investigate building a single datastore that includes all domains.
These test-time datasets can be either in-domain or out-of-domain with respect to the data used to train model parameters.

\subsection{
    Implementation Details
}\label{subsec:exp-setup}

\paragraph{LM architecture and training details.}
We use 1.3B-parameter transformer LMs based on the LLaMA architecture~\citep{touvron2023llama} as implemented in OpenLM.\footnote{\url{https://github.com/mlfoundations/openlm}}
Each model is trained with 128 A100 GPUs across 16 nodes. Following \citet{muennighoff2023scaling}, we train for multiple epochs in each dataset and perform early stopping. We train our \PD, \PD\SW\ and \PD\SW\BY~models for 60B, 250B, and 350B tokens in total, respectively. 
More details are provided in Appendix~\ref{app:model-details}.

\tightparagraph{Domain re-weighting.}
Since the distribution of \datanameshort~is highly skewed
(\S\ref{subsec:our-data-analysis}), we perform a simple upweighting scheme where we upsample all data 
that accounts for less than 5\%
by a factor of 3$\times$, which we found to work well after a sweep of different settings.
More sophisticated domain weighting strategies~\citep{xie2023doremi} are of interest but beyond the scope of this work.

\tightparagraph{Evaluation.} We benchmark our models using language modeling perplexity on 14 domains that represent both in-domain and out-of-domain data with respect to different levels of \datanameshort. This includes:
public-domain legal documents from the \textbf{FreeLaw} Project subset of the the Pile~\citep{gao2020pile}, 
a held-out collection of books from the \textbf{Gutenberg} collection~\citep{gutenberg},
conversational text from the \textbf{Hacker News} subset of the Pile,
held-out code files from the \textbf{Github} subset of the Pile (most of which are non-permissive licensed), scientific text of NIH Grant abstracts that are taken from the \textbf{NIH ExPorter} subset of the PILE, 
philosophy papers taken from the \textbf{PhilPapers} of the PILE,
held-out English \textbf{Wikipedia} articles from the PILE,
news articles from \textbf{CC-News}~\citep{mackenzie2020cc},
books from \textbf{BookCorpus2} which is an expanded version of \citet{zhu2015aligning},
books from \textbf{Books3} by \citet{presser2020books},
random web-crawled pages from \textbf{OpenWebText2}~\citep{openwebtext,gao2020pile}, emails from 
the \textbf{Enron Emails} corpus~\citep{klimt2004enron}, \textbf{Amazon} product reviews from \citet{he2016ups},
and finally clinical notes from \textbf{MIMIC-III}~\citep{johnson2016mimic} with personal identifiable information (PII) masked out.
Our choice of domains highlights legal risks discussed in the earlier sections, e.g., CC-News, BookCorpus2, Books3 and Amazon reviews are mostly copyrighted, Github is mostly not permissively licensed,\footnote{
    \citet{kocetkov2023the} estimates about 13\% of the Github data is under MIT, Apache, and BSD.
} and Enron Emails and MIMIC-III include private text.
We merge all text into one stream of text and split them into batches with a maximum sequence length of 1,024 and a sliding window of 512, a setup that is standard in prior language modeling literature~\citep{baevski2018adaptive,khandelwal2020generalization}.
For MIMIC-III, which includes masked personally-identifiable information (PII), we filter out notes where more than 50\% of tokens correspond to PII, and then exclude tokens corresponding to PII when computing perplexity.

\tightparagraph{Datastore.}
We construct an in-domain datastore for each test domain based on their training data. 
For datasets from the PILE, we consider 10\% of the training data. 
For \textbf{\knnlm}, each datastore consists of up to 1 billion $h$-dimensional vectors ($h=$2,048). We build an index for fast nearest neighbor search using FAISS~\citep{johnson2019billion}.
For \textbf{\riclm}, each datastore consists of text blocks with a length of 1,024 and a sliding window of 512.
We use BM25 from Pyserini~\citep{lin2021pyserini}.
Appendix~\ref{app:additional-nonparametric} report ablations on different implementations of \riclm\ besides the method in \S\ref{subsec:non-parametric-lm}. 
More details, statistics and hyperparameter values for the datastores are reported in \S\ref{app:model-details}.

\section{Experimental Results}\label{sec:exp}\sisetup{round-mode=places, round-precision=1}

\definecolor{orange}{HTML}{E66100}
\definecolor{bluex}{HTML}{0C7BDC}
\definecolor{yellow}{HTML}{FFC20A}

\newcommand{\ID}[1]{\cellcolor{bluex!25} \text{{#1}}}
\newcommand{\SD}[1]{\cellcolor{yellow!25} \text{{#1}}}
\newcommand{\OD}[1]{\cellcolor{orange!25} \text{{#1}}}
\newcommand{\SDRel}[1]{\cellcolor{yellow!25} $_\text{(-#1\%)}$}
\newcommand{\ODRel}[1]{\cellcolor{orange!25} $_\text{(-#1\%)}$}
\newcommand{\rel}[1]{$_\text{(-#1\%)}$}

\begin{table}[t]
    \small 
    \centering
    \begin{tabular}{l R{1.5cm} R{1.5cm} R{1.5cm} R{1.5cm}}
        \toprule
            Eval data & \PD & \PD\SW & \PD\SW\BY & Pythia \\
        \midrule
            FreeLaw         & \ID{5.3} & \ID{5.7} & \ID{6.5} & \ID{5.6} \\
            Gutenberg       & \ID{15.2} & \ID{12.5} & \ID{14.1} & \ID{13.1} \\
            HackerNews      & \OD{38.0} & \ID{13.7} & \ID{14.5} & \ID{13.3} \\
            Github          & \OD{13.5} & \SD{2.7} & \SD{2.8} & \ID{2.4} \\
            NIH ExPorter    & \OD{28.2} & \SD{19.2} & \SD{15.0} & \ID{11.1} \\
            PhilPapers      & \OD{31.7} & \SD{17.6} & \SD{15.0} & \ID{12.7} \\
            Wikipedia       & \OD{28.9} & \OD{20.3} & \ID{11.3} & \ID{9.1} \\
            CC News         & \OD{34.0} & \OD{23.3} & \OD{21.2} & \ID{12.0} \\
            BookCorpus2     & \OD{25.3} & \OD{19.2} & \OD{19.6} & \ID{13.2} \\
            Books3          & \OD{27.2} & \OD{19.3} & \OD{18.6} & \ID{12.6} \\
            OpenWebText2    & \OD{37.8} & \OD{21.1} & \OD{18.8} & \ID{11.5} \\
            Enron Emails    & \OD{18.6} & \OD{13.2} & \OD{13.5} & \ID{6.9} \\
            Amazon          & \OD{81.1} & \OD{34.8} & \OD{37.0} & \ID{22.9} \\
            MIMIC-III       & \OD{22.3} & \OD{19.0} & \OD{15.5} & \OD{13.1} \\
        \midrule
            Average         & 29.1 & 17.3 & 16.0 & 11.4 \\
        \bottomrule
    \end{tabular}
    \caption{
        Perplexity (the lower the better) of the parametric-only \modelname\ trained on \PD{}, \PD{}\SW{}, and \PD{}\SW{}\BY{} (without a datastore), compared to Pythia-1.4B, a model trained with similar amounts of compute but on mostly non-permissive data.
        We use \textcolor{bluex!25}{$\blacksquare$},
        \textcolor{orange!25}{$\blacksquare$},
        and \textcolor{yellow!25}{$\blacksquare$} to indicate text that is in-domain, out-of-domain, or out-of-domain but has relevant data in-domain (e.g., high-risk Github code vs. our permissive Github code). 
        Reported on the test data; see Table~\ref{tab:results-parameteric-val} for results on the validation data.
        \textbf{Our parametric LMs are competitive to Pythia in-domain but fall short out-of-domain.} 
    }\label{tab:results-parameteric}
\end{table}


We first evaluate the parametric-only component of \modelname~trained on the \dataname~(\S\ref{subsec:results-parametric}), and then show the effect of adding a datastore that may contain high-risk text (\S\ref{subsec:results-nonparametric}).
For all experiments, we use the 1.4B Pythia model~\citep{biderman2023pythia} as a baseline because it is trained with a similar amount of compute (data size and model parameters), but is trained on mostly high-risk data.\footnote{We use the model checkpoint from \url{https://huggingface.co/EleutherAI/pythia-1.4b-deduped-v0}.}

\subsection{Results: 
    Parametric Component
}\label{subsec:results-parametric}

\paragraph{Main results.}
Table~\ref{tab:results-parameteric} reports performance of our 1.3B base LMs trained on varying levels of permissively-licensed data---\PD, \PD\SW, and \PD\SW\BY---as well as Pythia.
Overall, our LMs are competitive with Pythia despite using permissive data only.
They are roughly equal quality on in-domain data, e.g., FreeLaw and Gutenberg, HackerNews in the case of \PD\SW\ and \PD\SW\BY, and Wikipedia in the case of \PD\SW\BY.
Models trained on \PD\SW\ and \PD\SW\BY\ are also close to Pythia on Github, likely because the permissively-licensed code data included in \SW\ has a distribution that is sufficiently close to the distribution of the all Github code.
The largest gaps occur on data that is in-domain for Pythia but out-of-domain for our model, e.g., news, books, OpenWebText, and emails, and Wikipedia in the case of models besides \PD\SW\BY.
This illustrates 
the extreme domain generalization challenge that is present when training on only permissive data, as we hint 
in \S\ref{subsec:our-data-analysis}.

\sisetup{round-mode=places, round-precision=1}

\tightparagraph{Gaps from Pythia align with a degree of domain shift.} 
The similarity of an evaluation domain to a domain of the \datanameshort\ strongly correlates with the performance gaps between \modelname~and Pythia. To show this, we compute the Pearson correlation between 1) the maximum $n$-gram overlap between an \datanameshort~domain and the Pile validation domains (from \S\ref{subsec:our-data-analysis}) and 2) the perplexity difference between the Pythia model and our \PD\SW~model, normalized by the performance of the \PD\SW~model. We find a strong negative correlation between these metrics ($r$=-0.72, $p <$ 0.005), indeed indicating that the more dissimilar an evaluation domain is from the \datanameshort~domains, the better Pythia does relative to \modelname\ (see \S\ref{app:additional-data-analysis} for a scatter plot). 

More ablations, including the effect of upsampling low-resource data, and the effect of including and excluding explicit source code, are provided in \S\ref{app:additional-parametric}.



\subsection{Results: Adding the Nonparametric Component}\label{subsec:results-nonparametric}

Since building legally permissive LMs poses a challenge of  extreme domain generalization, 
our next question is whether using an in-domain, nonparametric datastore can reduce the gap. 
We explore this question with our parametric LM trained on the \PD\SW~subset of \datanameshort; see Appendix~\ref{app:additional-nonparametric} for results of models trained on \PD\ or \PD\SW\BY. All models are evaluated on a subset of 8 out-of-domain datasets to the parametric model: Github, NIH ExPorter, Wikipedia, CC News, Books3, Enron Emails, Amazon, and MIMIC-III. 
\begin{table}[t]
    \small 
    \centering
    \setlength{\tabcolsep}{0.05cm}
    \begin{tabular}{
        l R{1.5cm} R{1.3cm} p{0.7cm} R{1.3cm} p{0.7cm} R{1.5cm}
    } 
        \toprule
            \multirow{2}{*}{Eval data} & \multicolumn{5}{c}{\modelname~(\PD\SW)} & \multicolumn{1}{c}{Pythia} \\
            \cmidrule(lr){2-6} \cmidrule(lr){7-7}
            & Prm-only &  \multicolumn{2}{r}{\knnlm~~} & \multicolumn{2}{r}{\riclm~~~} & Prm-only  \\
        \midrule
            Github          & \SD{2.7}  & \SD{2.4} & \SDRel{100} & \SD{2.4}  & \SDRel{100}& \ID{2.4}  \\
            NIH ExPorter~~~~~~~& \SD{19.2} & \SD{15.0}  & \SDRel{52} & \SD{18.5}  & \SDRel{9} & \ID{11.1} \\
            Wikipedia       & \OD{20.3} & \OD{14.5}  & \ODRel{52} & \OD{19.4} & \ODRel{8}& \ID{9.1} \\
            CC News         & \OD{23.3} & \OD{8.0} & \ODRel{135} & \OD{16.8} & \ODRel{58}& \ID{12.0} \\
            Books3          & \OD{19.3} & \OD{17.4} & \ODRel{28} & \OD{18.6} & \ODRel{10}& \ID{12.6} \\
            Enron Emails    & \OD{13.2} & \OD{5.9} & \ODRel{116}& \OD{9.9} & \ODRel{68}& \ID{6.9} \\
            Amazon          & \OD{34.9} & \OD{26.0} & \ODRel{75} & \OD{33.7} & \ODRel{10}& \ID{23.0} \\
            MIMIC-III       & \OD{19.0} & \OD{6.6} & \ODRel{210} & \OD{15.6} & \ODRel{58}& \OD{13.1} \\
        \midrule
            Average         & 19.0 & 12.0 & \rel{91} & 16.9 & \rel{27} & 11.3 \\
        \bottomrule
    \end{tabular}
    \caption{
        Perplexity (the lower the better) of parametric LMs (Prm-only), \knnlm, and \riclm.
        \% in parentheses indicate a reduction in the gap between the parametric-only \modelname\ and Pythia.
        As in Table~\ref{tab:results-parameteric}, \textcolor{bluex!25}{$\blacksquare$}
        indicates in-domain;
        \textcolor{orange!25}{$\blacksquare$}
        indicates out-of-domain;
        \textcolor{yellow!25}{$\blacksquare$} indicates out-of-domain but has relevant data in-domain, all with respect to the training data of the parametric LM. 
        Reported on the test data; see Table~\ref{tab:results-nonparameteric-val} for results on the validation data.
        See Table~\ref{tab:datastore-statistics} for the statistics of the datastore.
        \textbf{Adding a datastore, with \knnlmbold, 
        effectively reduces the gap between \modelname\ and Pythia. 
        } 
    }\label{tab:results-nonparameteric}
\end{table}

\begin{figure*}[t]
\centering
\includegraphics[trim={0cm, 1.5cm, 0.3cm, 0},clip,width=0.343\columnwidth]{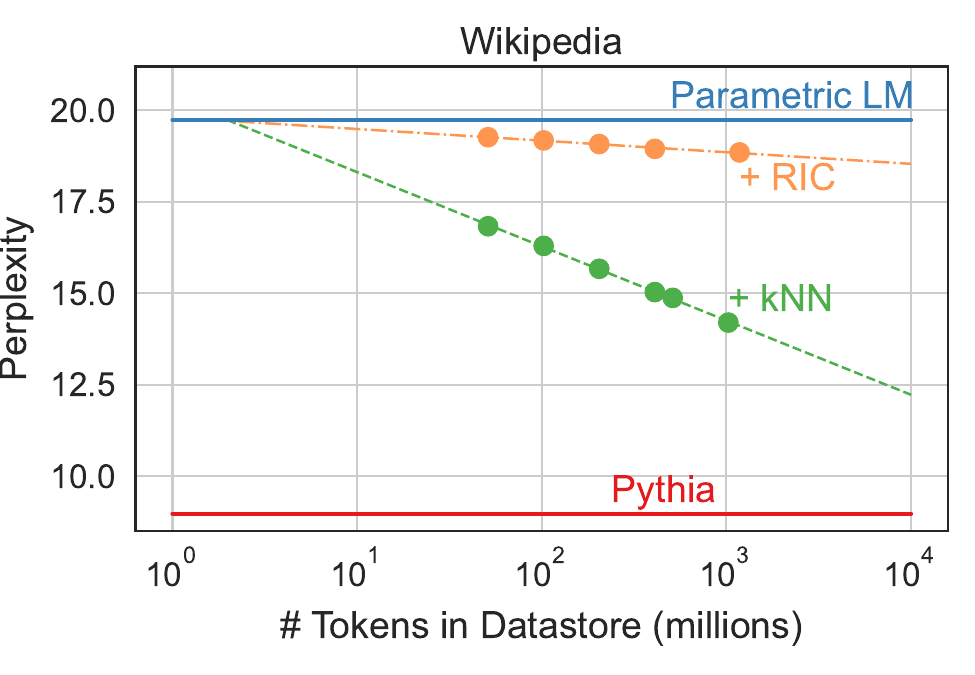}
\includegraphics[trim={1.1cm, 1.5cm, 0.3cm, 0},clip,width=0.32\columnwidth]{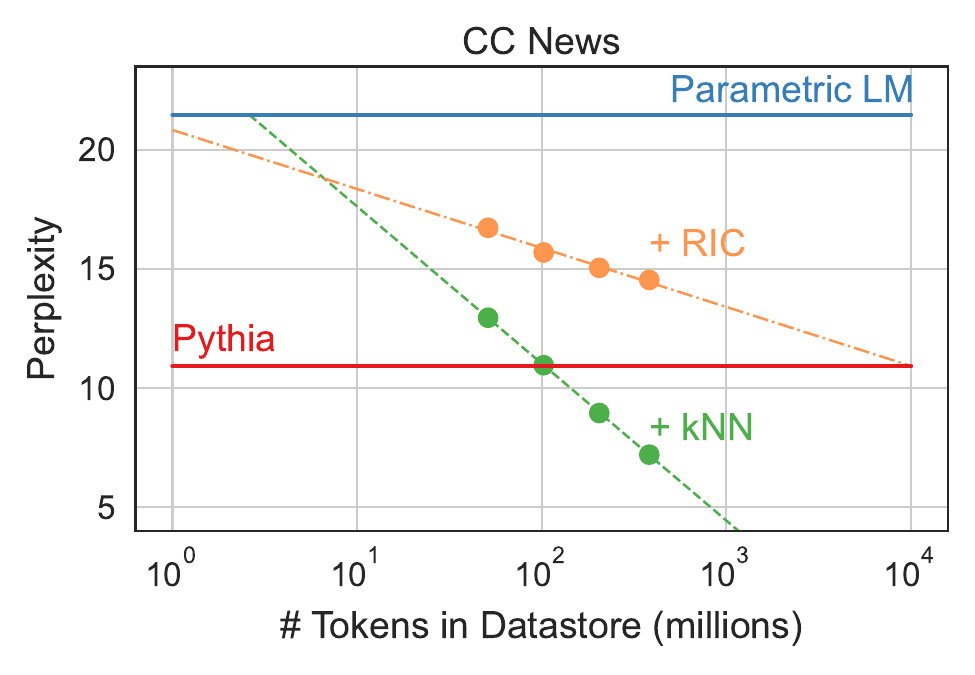}
\includegraphics[trim={1.1cm, 1.5cm, 0.3cm, 0},clip,width=0.32\columnwidth]{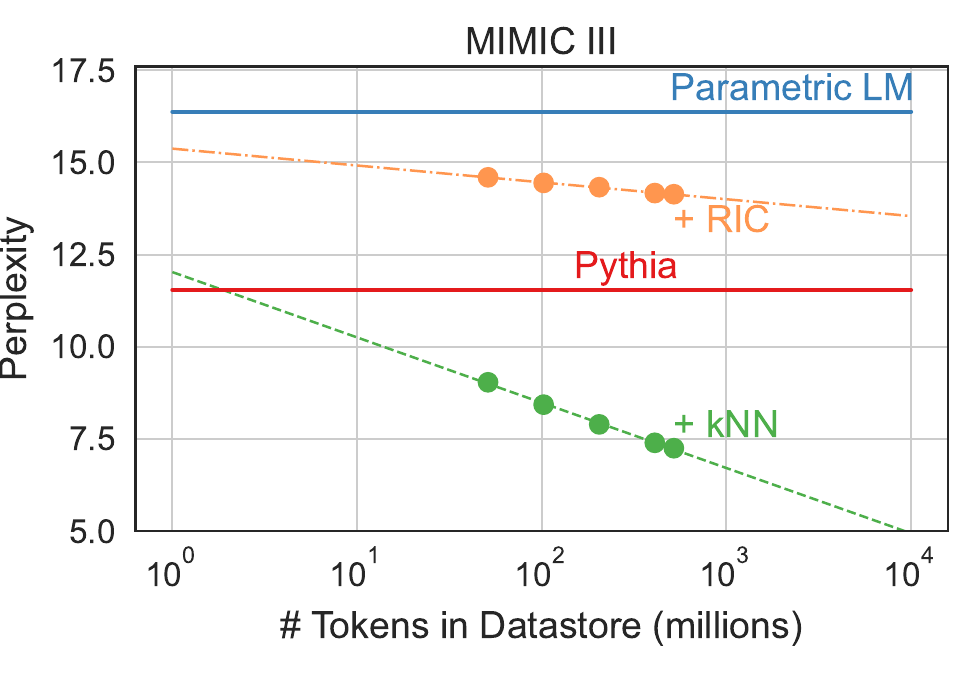}
\includegraphics[trim={0cm, 0.5cm, 0.3cm, 0},clip,width=0.343\columnwidth]{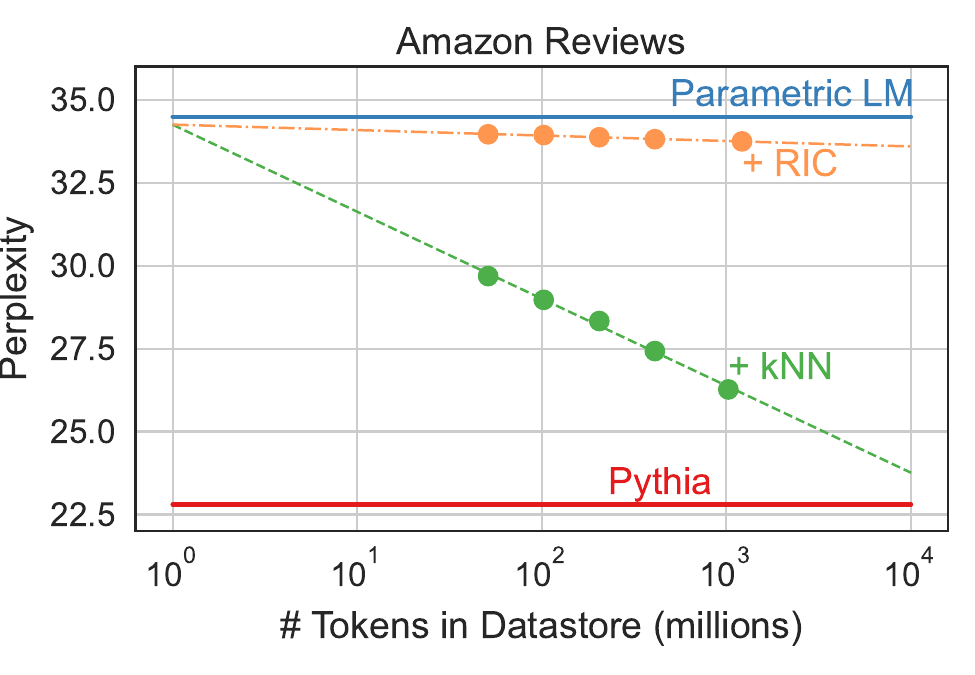}
\includegraphics[trim={1.1cm, 0.5cm, 0.3cm, 0},clip,width=0.32\columnwidth]{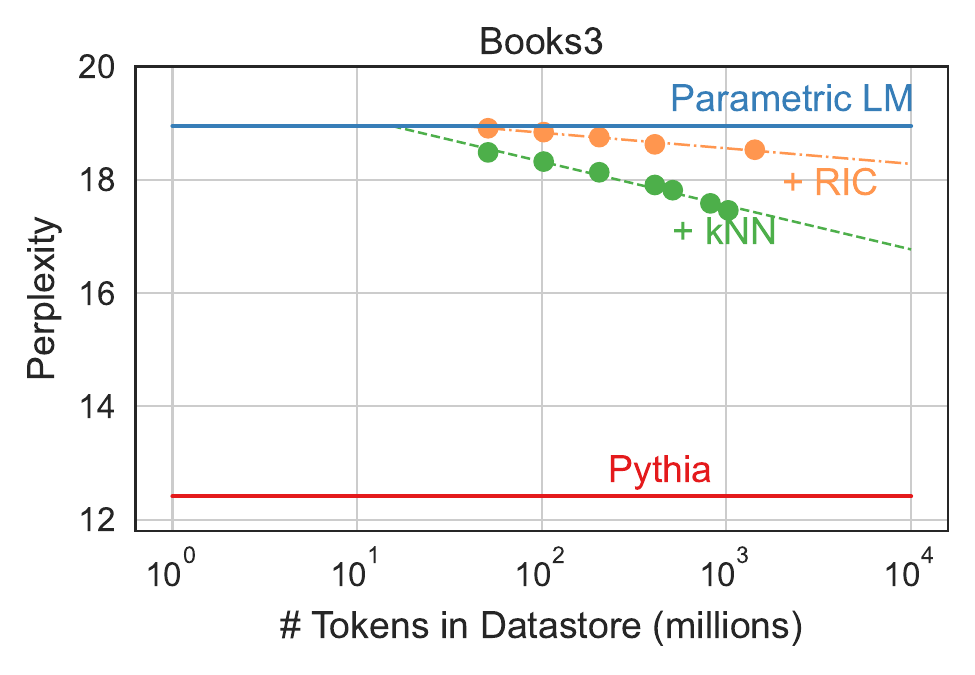}
\includegraphics[trim={1.1cm, 0.5cm, 0.3cm, 0},clip,width=0.32\columnwidth]{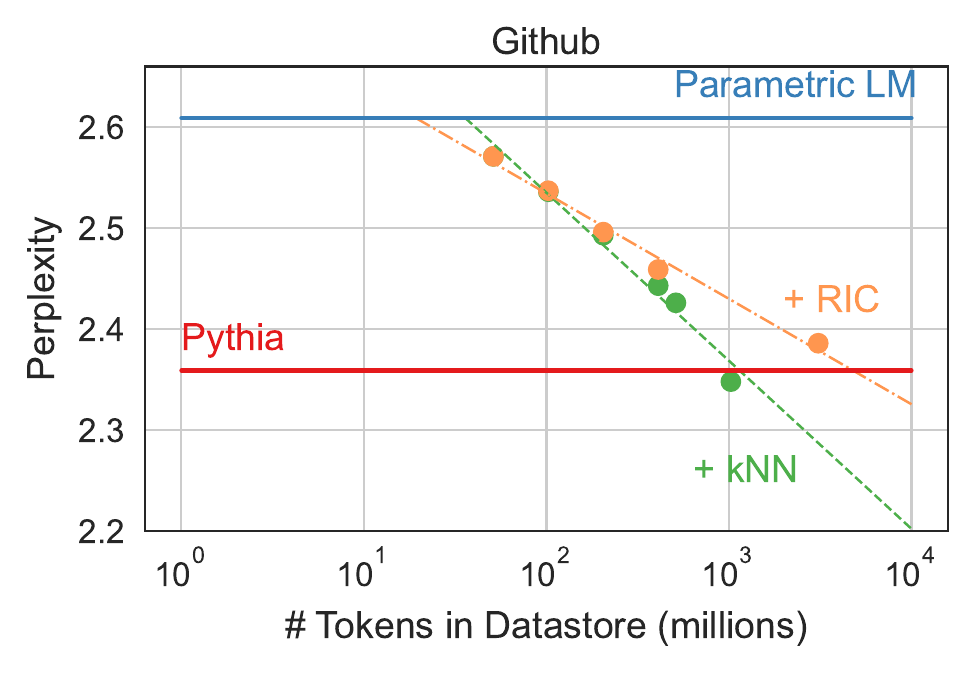}
\vspace{-.5em}
\caption{
    Impact of scaling the datastore of \modelname{}~(\PD\SW).
    Perplexity on random 128K tokens from the validation data reported.
    The rightmost dots for \knnlm\ and \riclm\ in each figure correspond to the final models used in Table~\ref{tab:results-nonparameteric}.
    \textbf{Scaling the test-time datastore consistently improves performance over all domains.} 
}
\label{fig:scale-datastore}
\end{figure*}


\tightparagraph{Main results.}
Table~\ref{tab:results-nonparameteric} shows adding the datastore with either \knnlm- or \riclm-based retrieval improves performance over just using the parameteric component on all domains, but
\knnlm\ is more effective than \riclm.
In most domains, \knnlm\ reduces the gap between \modelname\ and Pythia by more than 50\%\ (on NIH ExPorter, Wikipedia, Amazon) or even outperforms Pythia (on Github, CC News, Enron Emails, MIMIC-III).
Books3 is the domain with the least benefit, on which \knnlm\ still reduces the gap by 28\%. 

\tightparagraph{Impact of scaling the datastore.}
Figure \ref{fig:scale-datastore} demonstrates that both \knnlm~and \riclm-based retrieval consistently improves performance as the datastore size increases, with a strong log-linear trend. However, \knnlm\ improves performance more rapidly than \riclm\ does, consistently over all datasets.
%
Extrapolating the trend suggests that, on the domains that \modelname\ has not outperformed Pythia yet, scaling the datastore even further (with \knnlm~retrieval) may enable it to match Pythia. 


\tightparagraph{Why does \knnlmbold\ outperform \riclm?} Our next question is why \knnlm\ is better than \riclm---is it (a) because \knnlm\ is better than \riclm\ in general, 
or (b) because \knnlm\ 
generalizes out-of-domain better than \riclm\ does?
Our further analysis in \S\ref{app:additional-nonparametric} (Figure~\ref{fig:scale-datastore-all}) reveals that it is both.
With Pythia, where the test data is in-domain, while both \knnlm\ and \riclm\ improve performance upon the parametric-only model, \knnlm\ is overall better and scales better than \riclm, supporting (a).
Both \knnlm\ and \riclm\ improve performance more rapidly with \modelname\ (where the test data is out-of-domain) than with Pythia, but this trend is much clearer with \knnlm, supporting (b).

\begin{wrapfigure}{rt}{0.4\textwidth}
    \vspace{-0.1em}
    \captionof{table}{
    \textbf{Zero-shot performance on ten classification datasets.}
    \knnlm\ allows performance of \modelname\ to match that of Pythia by using a datastore.
    \textcolor{bluex!25}{$\blacksquare$} indicates in-domain; \textcolor{orange!25}{$\blacksquare$} indicates out-of-domain.
  } \vspace{-0.2em}
  \begin{center} \myfontsize \setlength{\tabcolsep}{4pt}
    \begin{tabular}{lrrr}
        \toprule
            \multirow{3}{*}{Eval} & \multicolumn{2}{c}{\modelname~(\PD\SW)} & \multicolumn{1}{c}{Pythia} \\
            \cmidrule(lr){2-3} \cmidrule(lr){4-4}
            & Prm-only & \knnlm & Prm-only \\
        \midrule
        AGN     & \OD{63.3} & \OD{79.5} & \ID{71.2} \\
        Dbpedia & \OD{36.9} & \OD{41.3} & \ID{39.1} \\
        SST-2   & \OD{57.1} & \OD{74.4} & \ID{79.7} \\
        MR      & \OD{55.1} & \OD{79.0}   & \ID{79.9} \\
        RT      & \OD{55.3} & \OD{67.9} & \ID{80.4} \\
        CR      & \OD{64.6} & \OD{83.1} & \ID{80.3} \\
        Yelp    & \OD{62.8} & \OD{84.5} & \ID{84.7} \\
        Amz     & \OD{60.4} & \OD{82.9} & \ID{80.3} \\
        RTE     & \OD{56.0} & \OD{55.2} & \ID{53.7} \\
        HYP     & \OD{58.5} & \OD{63.1} & \ID{58.5} \\

        \midrule
        Avg     & \OD{57.0} & \OD{71.1} & \ID{70.8} \\

        \bottomrule
    \end{tabular}
    \hfill \scriptsize
  \end{center}
  \label{tab:downstream}
  \vspace{-1em}
\end{wrapfigure}

\tightparagraph{Downstream task performance.}
In order to verify if our findings on language modeling perplexity transfer to downstream tasks, we evaluate zero-shot performance of \modelname\ and Pythia on ten text classification datasets whose domains are not covered by \datanameshort. All models use PMI~\citep{holtzman-etal-2021-surface} for a better calibration of model outputs, and we use $k$NN-Prompt~\citep{shi2022nearest} for applying \knnlm\ for downstream tasks. See \S\ref{app:model-details} for the details.
Table~\ref{tab:downstream} demonstrates that our earlier findings hold on all ten datasets: the parametric-only \modelname\ largely underperforms Pythia; however, adding a datastore greatly improves performance, allowing performance of \modelname\ to match that of Pythia.

\tightparagraph{Where does the remaining gap come from?}
%
Even when scaling the datastore with \knnlm, \modelname\ lags behind Pythia on a few domains. Moreover, a Pythia-based \knnlm\ outperforms our model since \knnlm\ improves Pythia as well.
There are two possible points of failure in our model for these cases: either the parametric component (which outputs $P_\mathrm{LM}$) struggles out-of-domain, or the encoder (that outputs $P_{k\mathrm{NN}}$) struggles out-of-domain. To better understand which part of the model contributes to the gap we observe, we vary \modelname~with different choices for the parametric component and the encoder. We compare replacing either the parametric component or the encoder with Pythia. This setup allows us to measure the effects of the out-of-domain nature of our parametric component (which is only trained on \PD\SW~subset of \datanameshort) in each of these components.

\begin{figure*}[t]
\centering
    \includegraphics[trim={0.4cm, 1.2cm, 0.3cm, 0.5cm},clip,width=0.7\columnwidth]{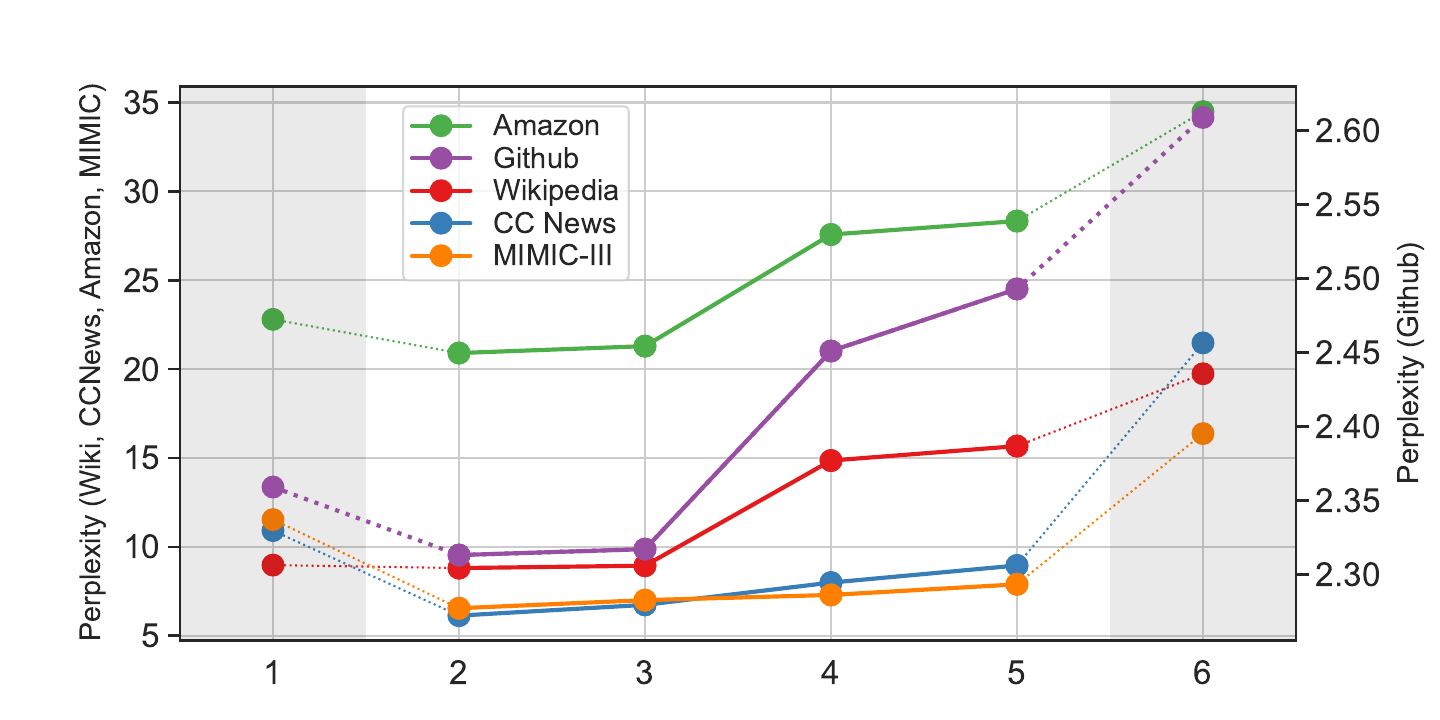}
    \vspace{0.06cm}
    \scriptsize
    \begin{tabular}{ R{2cm} P{0.88cm} P{0.88cm} P{0.88cm} P{0.88cm} P{0.88cm} P{0.88cm} P{1.7cm} }
        LM for $P_\mathrm{LM}$          & Pythia & Pythia & Pythia & Ours & Ours & Ours & \\[0.7ex]
        Encoder for $P_{k\mathrm{NN}}$  & \xmark & Pythia & Ours & Pythia & Ours & \xmark & \\
    \end{tabular}
\vspace{-.1em}
\caption{
        Impact of using different parameters on \modelname{}. Perplexity on random 128K tokens from the validation data reported. The left-most and the right-most models are parametric models, and the other four models are \knnlm{}s, using a datastore with 204.8 million tokens (20\% of the datastore we use for the main experiments). {\em Ours} indicates our parametric model trained on the \PD\SW~subset of \dataname.
        \textbf{Most of the performance degradation comes from using the out-of-domain parametric LM, rather than using the out-of-domain encoder.}
}
\label{fig:gap-analysis}
\end{figure*}

Results in Figure~\ref{fig:gap-analysis} reveal that most performance gaps come from the LM: performance improves significantly when the parametric component is replaced with Pythia, given a fixed encoder.
In contrast, performance improvement is relatively marginal when the encoder is replaced with Pythia, given a fixed parametric component.
These results indicate that the parametric component, which gives $P_\mathrm{LM}$, is quite sensitive to domain shift, but the encoder, which provides the nonparametric distribution $P_{k\mathrm{NN}}$, is fairly robust to extreme domain shift.
This also explains why \knnlm\ generalizes better than \riclm, since \riclm~is bottlenecked by the parametric component.

In summary, our analysis highlights two promising directions to further reduce the gap:\begin{enumerate}[itemsep=2pt, leftmargin=16pt, topsep=0pt]
    \item Scaling the datastore beyond 1 billion tokens, e.g., at the scale of trillions of tokens as in \citet{borgeaud2022improving}, as demonstrated by Figure~\ref{fig:scale-datastore}.
    \item Improving the robustness of the model by improving nonparametric techniques or designing a model that only uses a nonparametric distribution~\citep{min2022nonparametric}, as demonstrated by Figure~\ref{fig:gap-analysis}.
\end{enumerate}




\tightparagraph{Comparison in runtime speed.}
Table~\ref{tab:runtime} in Appendix~\ref{app:additional-nonparametric}  provides a comparison of the runtime speed of the parametric LM, \riclm, and \knnlm.
There is a strong tradeoff between performance and speed:
both \riclm\ and \knnlm\ are considerably slower than the parametric LM, and a larger datastore and more accurate nearest-neighbor search leads to better performance and slower inference.
While the speed is heavily influenced by the hardware used for benchmarking and thus it is difficult to precisely quantify how much faster one method is compared to the other, this suggests that improving the runtime efficiency of nonparametric approaches is an important area of future work. 


\subsection{Examples of Data Attribution and Opt-Out}
\label{subsec:attribution_optout}

\begin{wrapfigure}{rt}{0.45\textwidth}
  \begin{center} \myfontsize \setlength{\tabcolsep}{4pt}
    \vspace{0em}
    \begin{tabular}{lrrrr}
        \toprule
            \multirow{3}{*}{Eval} & \multicolumn{3}{c}{\modelname~(\PD\SW)} & \multicolumn{1}{c}{Pythia} \\
            \cmidrule(lr){2-4} \cmidrule(lr){5-5}
            & \multirow{2}{*}{Prm-only} & \knnlm\ & \knnlm\ & \multirow{2}{*}{Prm-only} \\
            && w/o HP & w/ HP \\
        \midrule
            1 & \OD{15.9} & \OD{15.2} & \OD{13.0} & \ID{9.6} \\
            2 & \OD{17.7} & \OD{16.7} & \OD{12.4} & \ID{10.0} \\
            3 & \OD{16.5} & \OD{15.6} & \OD{11.4} & \ID{9.5} \\
            4 & \OD{17.7} & \OD{16.8} & \OD{12.9} & \ID{10.1} \\
            5 & \OD{17.8} & \OD{16.9} & \OD{13.2} & \ID{10.2} \\
            6 & \OD{17.4} & \OD{16.5} & \OD{12.8} & \ID{10.1} \\
            7 & \OD{18.8} & \OD{17.8} & \OD{15.1} & \ID{10.9}  \\
        \midrule
            Avg & \OD{17.4} & \OD{16.5} & \OD{12.9} & \ID{10.1}  \\
        \bottomrule
    \end{tabular}
    \hfill \scriptsize
  \end{center}
  \vspace{-0.5em}
  \captionof{table}{
    \textbf{The effect of data opt-out.}
    Both \knnlm\ methods use 1.024B-token on Books3. {\em w/ HP} and {\em w/o HP} indicate that the datastore includes or excludes Harry Potter books, respectively. The number (1 to 7) indicates a different book from the Harry Potter series used as the eval data; this eval book is not included in the datastore in any case. \textcolor{bluex!25}{$\blacksquare$} indicates in-domain; \textcolor{orange!25}{$\blacksquare$} indicates out-of-domain.
  } \vspace{-2.0em}
  \label{tab:opt-out}
\end{wrapfigure}

As discussed in \S\ref{sec:background}, the design of \modelname\ can better align with various data-use regulations 
by providing mechanisms for data attribution during inference and for data owners to remove their data from the model at any time. This section show examples of such capabilities.

\tightparagraph{Data opt-out.}
To showcase the impact of opt-out on model performance, we conduct experiments with J.K.~Rowling's Harry Potter series. We first identify all seven Harry Potter books from the Books3 corpus of the Pile.
For each book, we calculate the perplexity of \modelname\ using two 1.024B token datastores on Books3, but one including the remaining six Harry Potter books and the other excluding any Harry Potter books.
This experiment is to see whether excluding Harry Potter books from the former datastore can reduce the likelihood of generating the leave-out Harry Potter book.

Table~\ref{tab:opt-out} shows the results. \modelname\ with Harry Potter books in the datastore effectively improves perplexity over all seven books, closing the gap between the \PD\SW\ model and Pythia.
However, when the Harry Potter books are removed from the datastore, the perplexity gets worse, approaching that of the parametric-only LM. This illustrates that eliminating the effect of the Harry Potter books from the model substantially reduces the likelihood of generating the leave-out book.

\tightparagraph{Attribution examples.}
To show the attribution feature of our model, Table~\ref{tab:analysis} provides qualitative examples on the top-$1$ context retrieved by \modelname. 
The model is able to assign a high probability to the ground truth token by retrieving highly relevant context.
It achieves this by leveraging the unique characteristics of the text within the datastore, such as recognizing that {\em Azkaban} refers to the prison and {\em green light} is associated with the {\em Killing Curse} in the Harry Potter books.

More qualitative examples on Github, news and emails are illustrated in Table~\ref{tab:qualitative-examples} in Appendix~\ref{app:additional-nonparametric}. They highlight that a nonparametric approach addresses specific legal risks that we have discussed earlier, e.g., it offers per-token attribution for free, and can provide a copyright notice when part of copyrighted text is being used for the probability distribution.

\begin{table*}
\centering \myfontsize
    \begin{tabular}{p{13.5cm}}
        \toprule
            \testPrefix\ `I - what - {\em dragons}?' spluttered the Prime Minister. `Yes, three,' said Fudge. `And a sphinx. Well, good day to you.' The Prime Minister hoped beyond hope that dragons and sphinxes would be the worst of it, but no. Less than two years later, Fudge had erupted out of the fire yet again, this time with the news that there had been a mass breakout from \\
            \testContinuation\ \cn{Azkaban}. `A {\em mass} breakout?' the Prime Minister had repeated hoarsely. \\
            \retrievedPrefix\ `D' you know Crouch, then?' said Harry. 
            Sirius' face darkened. He suddenly looked as menacing as the night when Harry had first met him, the night when Harry had still believed Sirius to be a murderer.
            `Oh, I know Crouch all right,' he said quietly. `He was the one who gave me the order to be sent to \\
            \retrievedContinuation\ \cn{Azkaban} - without a trial.' \\
        \midrule
            \testPrefix\ Terror tore at Harry's heart... he had to get to Dumbledore and he had to catch Snape... somehow the two things were linked... he could reverse what had happened if he had them both together... Dumbledore could not have died... (...) Harry felt Greyback collapse against him; with a stupendous effort he pushed the werewolf off and onto the floor as a jet of \\
            \testContinuation\ \cn{green} light came flying toward him; he ducked and ran, headfirst, into the fight. \\
            \retrievedPrefix\ Voldemort was ready. As Harry shouted, ``Expelliarmus!'' Voldemort cried, ``Avada Kedavra!'' A jet of \\
            \retrievedContinuation\ \cn{green} light issued from Voldemort's wand just as a jet of red light blasted from Harry's ... \\
        \bottomrule
    \end{tabular}
    \caption{
    \textbf{Attribution examples on Harry Potter books.} We show the top-$1$ retrieved context of \modelname\ (\PD\SW).
    \cn{Red underline text} indicates the next token that immediately follows the prefix.
    In both examples, the test data is from the sixth novel and the retrieved context is from the fourth novel in the Harry Potter series.
    In the series, {\em Azkaban} is the notorious wizarding prison, and the {\em green light} is a distinct characteristic of the Killing Curse, {\em Avada Kedavra}.
    }\label{tab:analysis}
\end{table*}




\section{Discussion \& Future Work}\label{sec:discuss}

Our work suggests that it is possible to improve the tradeoff between legal risk and model performance when training LMs. Our approach provides new options for model designers to mitigate the legal risk of LMs, and empowers stakeholders to have more control over the data that drives these systems. We point out a number of rich areas for future work, beyond what was mentioned throughout the paper: 

\tightparagraph{Addressing the limitations of \modelname.}
\modelname\ does not completely eliminate legal risk. Instead, it provides users more control over the model's generated content and functionalities to better align with legal regulations.
For instance, \modelname\ does not remove the need for obtaining permission to use copyrighted content in a datastore when providing attribution is not sufficient, but its opt-out capabilities can strengthen fair use defense. Moreover, \modelname\ does not prevent copying copyright content from a datastore, but it offers 
a way to prevent generating sensitive text~\citep{huang2023privacy} or prevent copying the 
content verbatim.
These functionalities increase the likelihood of a successful fair use defense if used appropriately.

Furthermore, while \modelname\ mitigates copyright and privacy risks, it may exacerbate certain fairness issues, like toxicity towards marginalized groups and racial biases, especially due to the prevalence of older copyright-expired books in the training data. Exploring the balance between legal risk mitigation and fairness is an important future direction.

Finally, our study relies on explicit metadata to identify licenses, which may lead to underestimates of the amount and diversity of permissively licensed text actually available on the web. Future research may investigate \emph{inferring} data licenses from documents in web crawl at scale, which may be an effective way to build more heterogeneous, permissively licensed corpora.

\tightparagraph{Introducing novel data licensing approaches.}
\modelname\ introduces the possibility for data owners to set different levels of permissivity for learning parameters and for including in a nonparametric datastore. A data owner might choose to be more permissive about including data in the  datastore due to its ease of removal, ensuring that the excluded data has no influence on model predictions anymore, and its ability to provide per-prediction attribution.
Moreover, we envision that \modelname~could provide a path forward for data owners to get properly credited (or be paid directly) every time their data in a datastore contributes to a prediction. This is orthogonal to recent work that circumvented copyright issues by licensing out training data from data creators~\citep{meta2023}.

\tightparagraph{Investigating other copyright risk mitigation strategies.} It is critical to continue to develop new techniques that use copyrighted data while protecting the rights of data owners and subjects. In addition to nonparametric approaches, there are many other ways to achieve these goals. First, one could train LMs on copyrighted content but filter and guide their outputs towards text that is non-infringing~\citep{henderson2023foundation}. Second, training models with differential privacy~\citep{dwork2006calibrating,abadi2016deep} or near-access freeness~\citep{vyas2023provable} may prevent them from regenerating individual details of copyright data. Finally, one could provide attributions for standard base LMs using post-hoc attribution methods, e.g., influence functions~\citep{koh2017understanding}, rather than switching the model class to a retrieval-based model. All of these methods are complementary and orthogonal to our proposed approach.

\tightparagraph{Generalizing \modelname~as a modular language model.} Our work is closely related to recent studies on modular LMs, which have specialized parameters (or \emph{experts}) trained on different domains \citep{gururangan-etal-2022-demix, li2022branch, gururangan2023scaling}, languages \citep{pfeiffer2020madx, pfeiffer2022lifting}, or tasks \citep{chen2022modsquad, jang2023exploring}. Our work extends modular LMs to include nonparametric datastores, and focuses on {\em specializing} different parts of the model to low- and high-risk subsets of the training data. Legal risks may also be mitigated with a collection of parametric expert models that are specialized to low- and high-risk data. Future work may explore this possibility as well as the usefulness of combining a nonparametric datastore with parametric experts.

\tightparagraph{Extending \modelname~to other modalities.}
While this work focuses on text-only models,
similar methods to ours could apply to other domains and modalities. For instance, it might be possible to build permissive text-to-image generative models~\citep{rombach2022high} using compartmentalized public domain pre-training and retrieval-augmentation~\citep{chen2022re, golatkar2023training}. We believe such approaches are especially promising because there are many sources of public domain data in other modalities, e.g., images, speech, video, and more.
\section{Conclusion}\label{sec:concl}We introduce \modelname, a language model that mitigates legal risk by learning parameters only on low-risk, permissively-licensed data (\dataname), and using an unrestricted nonparametric datastore during inference. Our approach allows the model designer to use high-risk data without training on it, supports sentence-level data attribution, and enables data produces to opt-out from the model by removing content from the datastore.
Experiments on language modeling perplexity show that parametric-only \modelname\ is competitive on domains covered by \dataname, but falls short out-of-domain when solely using the parametric component of the model, highlighting the challenge of extreme domain generalization.
We then show that adding a nonparametric datastore to \modelname\ (with \knnlm\ retrieval) successfully addresses this challenge, significantly reducing the gap (or even outperforming) the Pythia baseline that is trained unrestrictedly. 
We show that scaling the datastore size is key to the success of the nonparametric approach, and that the encoder for a nonparametric distribution is significantly more robust to distribution shift than the parametric component. Our results point to a number of exciting future research directions to develop AI systems with mitigated legal risk.

\if\iclrsubmission1

\else
    \subsubsection*{Acknowledgments} We thank Peter Henderson and Mark Lemley for discussing the legality of LMs, and Kyle Lo for feedback on our dataset and license taxonomy.
    We thank Mitchell Wortsman for help with setting up compute and model training. We thank Chris Callison-Burch, James Grimmelmann, Tatsunori Hashimoto, Peter Henderson, Nikhil Kandpal, Pang Wei Koh, Mark Lemley, Kyle Lo, Fatemeh Mireshghallah, Sewoong Oh and Rulin Shao for valuable feedback on the project and the paper. We thank Matthijs Douze, Gergely Szilvasy, and Maria Lomeli for answering questions about FAISS, and Dirk Groeneveld for giving early access to the deduplication script.
    Sewon Min is supported by the J.P. Morgan Ph.D. Fellowship. Suchin Gururangan is supported by the Bloomberg Data Science Ph.D. Fellowship. Eric Wallace is supported by the Apple Scholars in AI/ML Fellowship. We thank Stability AI for providing compute to train the LMs in this work. At UW, this work supported in part by the Office of Naval Research under MURI grant N00014-18-1-2670, the DARPA MCS program through NIWC Pacific (N66001-19-2-4031), NSF IIS-20444660, and gifts from AI2.
\fi

\bibliography{iclr2024_conference}

\begin{thebibliography}{94}
\providecommand{\natexlab}[1]{#1}
\providecommand{\url}[1]{\texttt{#1}}
\expandafter\ifx\csname urlstyle\endcsname\relax
  \providecommand{\doi}[1]{doi: #1}\else
  \providecommand{\doi}{doi: \begingroup \urlstyle{rm}\Url}\fi

\bibitem[Abadi et~al.(2016)Abadi, Chu, Goodfellow, McMahan, Mironov, Talwar,
  and Zhang]{abadi2016deep}
Martin Abadi, Andy Chu, Ian Goodfellow, H~Brendan McMahan, Ilya Mironov, Kunal
  Talwar, and Li~Zhang.
\newblock Deep learning with differential privacy.
\newblock In \emph{ACM SIGSAC}, 2016.

\bibitem[ArXiv(2023)]{arxiv_dataset}
ArXiv.
\newblock arxiv dataset, 2023.
\newblock URL \url{https://www.kaggle.com/dsv/6015950}.

\bibitem[Baevski \& Auli(2019)Baevski and Auli]{baevski2018adaptive}
Alexei Baevski and Michael Auli.
\newblock Adaptive input representations for neural language modeling.
\newblock In \emph{Proceedings of the International Conference on Learning
  Representations}, 2019.

\bibitem[Bandy \& Vincent(2021)Bandy and Vincent]{bandy2021addressing}
Jack Bandy and Nicholas Vincent.
\newblock Addressing ``documentation debt''' in machine learning: A
  retrospective datasheet for {BookCorpus}.
\newblock In \emph{Proceedings of Advances in Neural Information Processing
  Systems}, 2021.

\bibitem[Biderman et~al.(2023)Biderman, Schoelkopf, Anthony, Bradley, O'Brien,
  Hallahan, Khan, Purohit, Prashanth, Raff, et~al.]{biderman2023pythia}
Stella Biderman, Hailey Schoelkopf, Quentin Anthony, Herbie Bradley, Kyle
  O'Brien, Eric Hallahan, Mohammad~Aflah Khan, Shivanshu Purohit, USVSN~Sai
  Prashanth, Edward Raff, et~al.
\newblock Pythia: A suite for analyzing large language models across training
  and scaling.
\newblock In \emph{Proceedings of the International Conference on Learning
  Representations}, 2023.

\bibitem[Black et~al.(2022)Black, Biderman, Hallahan, Anthony, Gao, Golding,
  He, Leahy, McDonell, Phang, Pieler, Prashanth, Purohit, Reynolds, Tow, Wang,
  and Weinbach]{black2022gptneox20b}
Sid Black, Stella Biderman, Eric Hallahan, Quentin Anthony, Leo Gao, Laurence
  Golding, Horace He, Connor Leahy, Kyle McDonell, Jason Phang, Michael Pieler,
  USVSN~Sai Prashanth, Shivanshu Purohit, Laria Reynolds, Jonathan Tow, Ben
  Wang, and Samuel Weinbach.
\newblock {GPT-NeoX-20B}: An open-source autoregressive language model.
\newblock \emph{arXiv preprint arXiv:2204.06745}, 2022.

\bibitem[Borgeaud et~al.(2022)Borgeaud, Mensch, Hoffmann, Cai, Rutherford,
  Millican, Van Den~Driessche, Lespiau, Damoc, Clark,
  et~al.]{borgeaud2022improving}
Sebastian Borgeaud, Arthur Mensch, Jordan Hoffmann, Trevor Cai, Eliza
  Rutherford, Katie Millican, George~Bm Van Den~Driessche, Jean-Baptiste
  Lespiau, Bogdan Damoc, Aidan Clark, et~al.
\newblock Improving language models by retrieving from trillions of tokens.
\newblock In \emph{Proceedings of the International Conference of Machine
  Learning}, 2022.

\bibitem[Bourtoule et~al.(2020)Bourtoule, Chandrasekaran, Choquette-Choo, Jia,
  Travers, Zhang, Lie, and Papernot]{bourtoule2020machine}
Lucas Bourtoule, Varun Chandrasekaran, Christopher~A. Choquette-Choo, Hengrui
  Jia, Adelin Travers, Baiwu Zhang, David Lie, and Nicolas Papernot.
\newblock Machine unlearning, 2020.

\bibitem[Brittain(2023)]{brittain2023copyright}
Blake Brittain.
\newblock {U.S.} copyright office says some {AI}-assisted works may be
  copyrighted.
\newblock \emph{Reuters}, 2023.

\bibitem[Brown et~al.(2020)Brown, Mann, Ryder, Subbiah, Kaplan, Dhariwal,
  Neelakantan, Shyam, et~al.]{gpt3}
Tom~B. Brown, Benjamin Mann, Nick Ryder, Melanie Subbiah, Jared Kaplan,
  Prafulla Dhariwal, Arvind Neelakantan, Pranav Shyam, et~al.
\newblock Language models are few-shot learners.
\newblock In \emph{NeurIPS}, 2020.

\bibitem[Cao \& Yang(2015)Cao and Yang]{cao2015towards}
Yinzhi Cao and Junfeng Yang.
\newblock Towards making systems forget with machine unlearning.
\newblock In \emph{2015 IEEE symposium on security and privacy}, pp.\
  463--480. IEEE, 2015.

\bibitem[Carlini et~al.(2021)Carlini, Tramer, Wallace, Jagielski, Herbert-Voss,
  Lee, Roberts, Brown, Song, Erlingsson, et~al.]{carlini2021extracting}
Nicholas Carlini, Florian Tramer, Eric Wallace, Matthew Jagielski, Ariel
  Herbert-Voss, Katherine Lee, Adam Roberts, Tom~B Brown, Dawn Song, Ulfar
  Erlingsson, et~al.
\newblock Extracting training data from large language models.
\newblock In \emph{USENIX Security Symposium}, 2021.

\bibitem[Carlini et~al.(2023)Carlini, Ippolito, Jagielski, Lee, Tramer, and
  Zhang]{carlini2022quantifying}
Nicholas Carlini, Daphne Ippolito, Matthew Jagielski, Katherine Lee, Florian
  Tramer, and Chiyuan Zhang.
\newblock Quantifying memorization across neural language models.
\newblock In \emph{Proceedings of the International Conference on Learning
  Representations}, 2023.

\bibitem[{Caselaw Access Project}()]{caselaw2018}
{Caselaw Access Project}.
\newblock Caselaw access project.
\newblock URL \url{https://case.law/}.

\bibitem[Chang et~al.(2023)Chang, Cramer, Soni, and Bamman]{chang2023speak}
Kent~K Chang, Mackenzie Cramer, Sandeep Soni, and David Bamman.
\newblock Speak, memory: An archaeology of books known to {ChatGPT}/{GPT}-4.
\newblock \emph{arXiv preprint arXiv:2305.00118}, 2023.

\bibitem[Chen et~al.(2022{\natexlab{a}})Chen, Hu, Saharia, and
  Cohen]{chen2022re}
Wenhu Chen, Hexiang Hu, Chitwan Saharia, and William~W Cohen.
\newblock {Re-Imagen}: Retrieval-augmented text-to-image generator.
\newblock In \emph{NeurIPS}, 2022{\natexlab{a}}.

\bibitem[Chen et~al.(2022{\natexlab{b}})Chen, Shen, Ding, Chen, Zhao,
  Learned-Miller, and Gan]{chen2022modsquad}
Zitian Chen, Yikang Shen, Mingyu Ding, Zhenfang Chen, Hengshuang Zhao, Erik
  Learned-Miller, and Chuang Gan.
\newblock Mod-squad: Designing mixture of experts as modular multi-task
  learners, 2022{\natexlab{b}}.

\bibitem[Dagan et~al.(2005)Dagan, Glickman, and Magnini]{dagan2005pascal}
Ido Dagan, Oren Glickman, and Bernardo Magnini.
\newblock The pascal recognising textual entailment challenge.
\newblock In \emph{Machine learning challenges workshop}, 2005.

\bibitem[De~Vynck(2023)]{de2023chatgpt}
Gerrit De~Vynck.
\newblock {ChatGPT} maker {OpenAI} faces a lawsuit over how it used people’s
  data, 2023.
\newblock URL
  \url{https://www.washingtonpost.com/technology/2023/06/28/openai-chatgpt-lawsuit-class-action/}.

\bibitem[Dodge et~al.(2021)Dodge, Sap, Marasovi{\'c}, Agnew, Ilharco,
  Groeneveld, Mitchell, and Gardner]{dodge2021documenting}
Jesse Dodge, Maarten Sap, Ana Marasovi{\'c}, William Agnew, Gabriel Ilharco,
  Dirk Groeneveld, Margaret Mitchell, and Matt Gardner.
\newblock Documenting large webtext corpora: A case study on the colossal clean
  crawled corpus.
\newblock In \emph{Proceedings of Empirical Methods in Natural Language
  Processing}, 2021.

\bibitem[Drozdov et~al.(2022)Drozdov, Wang, Rahimi, McCallum, Zamani, and
  Iyyer]{drozdov2022you}
Andrew Drozdov, Shufan Wang, Razieh Rahimi, Andrew McCallum, Hamed Zamani, and
  Mo~hit Iyyer.
\newblock You can't pick your neighbors, or can you? when and how to rely on
  retrieval in the {$k$NN-LM}.
\newblock \emph{arXiv preprint arXiv:2210.15859}, 2022.

\bibitem[Dwork et~al.(2006)Dwork, McSherry, Nissim, and
  Smith]{dwork2006calibrating}
Cynthia Dwork, Frank McSherry, Kobbi Nissim, and Adam Smith.
\newblock Calibrating noise to sensitivity in private data analysis.
\newblock In \emph{Theory of Cryptography}, 2006.

\bibitem[Fried et~al.(2023)Fried, Aghajanyan, Lin, Wang, Wallace, Shi, Zhong,
  Yih, Zettlemoyer, and Lewis]{fried2023incoder}
Daniel Fried, Armen Aghajanyan, Jessy Lin, Sida Wang, Eric Wallace, Freda Shi,
  Ruiqi Zhong, Scott Yih, Luke Zettlemoyer, and Mike Lewis.
\newblock Incoder: A generative model for code infilling and synthesis.
\newblock In \emph{Proceedings of the International Conference on Learning
  Representations}, 2023.

\bibitem[Gao et~al.(2020)Gao, Biderman, Black, Golding, Hoppe, Foster, Phang,
  He, Thite, Nabeshima, et~al.]{gao2020pile}
Leo Gao, Stella Biderman, Sid Black, Laurence Golding, Travis Hoppe, Charles
  Foster, Jason Phang, Horace He, Anish Thite, Noa Nabeshima, et~al.
\newblock The {Pile}: An {800GB} dataset of diverse text for language modeling.
\newblock \emph{arXiv preprint arXiv:2101.00027}, 2020.

\bibitem[Gershgorn(2021)]{gershgorn2021}
David Gershgorn.
\newblock Github’s automatic coding tool rests on untested legal ground,
  2021.
\newblock URL \url{https://www.theverge.com/
  2021/7/7/22561180/github-copilot-legal-copyright-fair-use-public-code}.

\bibitem[Gokaslan \& Cohen(2019)Gokaslan and Cohen]{openwebtext}
Aaron Gokaslan and Vanya Cohen.
\newblock {OpenWebText} corpus.
\newblock \url{http://Skylion007.github.io/}, 2019.

\bibitem[Golatkar et~al.(2023)Golatkar, Achille, Swaminathan, and
  Soatto]{golatkar2023training}
Aditya Golatkar, Alessandro Achille, Ashwin Swaminathan, and Stefano Soatto.
\newblock Training data protection with compositional diffusion models, 2023.

\bibitem[Groeneveld(2023)]{bff}
Dirk Groeneveld.
\newblock The big friendly filter.
\newblock \url{https://github.com/allenai/bff}, 2023.

\bibitem[Gururangan et~al.(2022)Gururangan, Lewis, Holtzman, Smith, and
  Zettlemoyer]{gururangan-etal-2022-demix}
Suchin Gururangan, Mike Lewis, Ari Holtzman, Noah~A. Smith, and Luke
  Zettlemoyer.
\newblock {DEM}ix layers: Disentangling domains for modular language modeling.
\newblock In \emph{Conference of the North American Chapter of the Association
  for Computational Linguistics}, 2022.

\bibitem[Gururangan et~al.(2023)Gururangan, Li, Lewis, Shi, Althoff, Smith, and
  Zettlemoyer]{gururangan2023scaling}
Suchin Gururangan, Margaret Li, Mike Lewis, Weijia Shi, Tim Althoff, Noah~A.
  Smith, and Luke Zettlemoyer.
\newblock Scaling expert language models with unsupervised domain discovery,
  2023.

\bibitem[Guu et~al.(2020)Guu, Lee, Tung, Pasupat, and Chang]{guu2020retrieval}
Kelvin Guu, Kenton Lee, Zora Tung, Panupong Pasupat, and Mingwei Chang.
\newblock Retrieval augmented language model pre-training.
\newblock In \emph{Proceedings of the International Conference of Machine
  Learning}, 2020.

\bibitem[Han et~al.(2023)Han, Simig, Mihaylov, Tsvetkov, Celikyilmaz, and
  Wang]{han2023understanding}
Xiaochuang Han, Daniel Simig, Todor Mihaylov, Yulia Tsvetkov, Asli Celikyilmaz,
  and Tianlu Wang.
\newblock Understanding in-context learning via supportive pretraining data.
\newblock In \emph{Proceedings of the Association for Computational
  Linguistics}, 2023.

\bibitem[He et~al.(2021)He, Neubig, and
  Berg-Kirkpatrick]{he-etal-2021-efficient}
Junxian He, Graham Neubig, and Taylor Berg-Kirkpatrick.
\newblock Efficient nearest neighbor language models.
\newblock In \emph{Proceedings of Empirical Methods in Natural Language
  Processing}, 2021.

\bibitem[He \& McAuley(2016)He and McAuley]{he2016ups}
Ruining He and Julian McAuley.
\newblock Ups and downs: Modeling the visual evolution of fashion trends with
  one-class collaborative filtering.
\newblock In \emph{Proceedings of the World Wide Web Conference}, 2016.

\bibitem[Henderson et~al.(2022)Henderson, Krass, Zheng, Guha, Manning,
  Jurafsky, and Ho]{henderson2022pile}
Peter Henderson, Mark Krass, Lucia Zheng, Neel Guha, Christopher~D Manning, Dan
  Jurafsky, and Daniel Ho.
\newblock Pile of {Law}: Learning responsible data filtering from the law and a
  {256GB} open-source legal dataset.
\newblock \emph{Proceedings of Advances in Neural Information Processing
  Systems}, 2022.

\bibitem[Henderson et~al.(2023)Henderson, Li, Jurafsky, Hashimoto, Lemley, and
  Liang]{henderson2023foundation}
Peter Henderson, Xuechen Li, Dan Jurafsky, Tatsunori Hashimoto, Mark~A Lemley,
  and Percy Liang.
\newblock Foundation models and fair use.
\newblock \emph{arXiv preprint arXiv:2303.15715}, 2023.

\bibitem[Hendrycks et~al.(2021)Hendrycks, Burns, Kadavath, Arora, Basart, Tang,
  Song, and Steinhardt]{hendrycksmath2021}
Dan Hendrycks, Collin Burns, Saurav Kadavath, Akul Arora, Steven Basart, Eric
  Tang, Dawn Song, and Jacob Steinhardt.
\newblock Measuring mathematical problem solving with the math dataset.
\newblock \emph{Proceedings of Advances in Neural Information Processing
  Systems}, 2021.

\bibitem[Holtzman et~al.(2021)Holtzman, West, Shwartz, Choi, and
  Zettlemoyer]{holtzman-etal-2021-surface}
Ari Holtzman, Peter West, Vered Shwartz, Yejin Choi, and Luke Zettlemoyer.
\newblock Surface form competition: Why the highest probability answer isn{'}t
  always right.
\newblock In \emph{Proceedings of Empirical Methods in Natural Language
  Processing}, 2021.

\bibitem[Hu \& Liu(2004)Hu and Liu]{hu2004mining}
Minqing Hu and Bing Liu.
\newblock Mining and summarizing customer reviews.
\newblock In \emph{Knowledge Discovery and Data Mining}, 2004.

\bibitem[Huang et~al.(2023{\natexlab{a}})Huang, Gupta, Zhong, Li, and
  Chen]{huang2023privacy}
Yangsibo Huang, Samyak Gupta, Zexuan Zhong, Kai Li, and Danqi Chen.
\newblock Privacy implications of retrieval-based language models.
\newblock \emph{arXiv preprint arXiv:2305.14888}, 2023{\natexlab{a}}.

\bibitem[Huang et~al.(2023{\natexlab{b}})Huang, Liu, Zhong, Shi, and
  Lee]{huang2023k}
Yangsibo Huang, Daogao Liu, Zexuan Zhong, Weijia Shi, and Yin~Tat Lee.
\newblock {$k$NN-Adapter}: Efficient domain adaptation for black-box language
  models.
\newblock \emph{arXiv preprint arXiv:2302.10879}, 2023{\natexlab{b}}.

\bibitem[Izacard et~al.(2022)Izacard, Lewis, Lomeli, Hosseini, Petroni, Schick,
  Dwivedi-Yu, Joulin, Riedel, and Grave]{izacard2022few}
Gautier Izacard, Patrick Lewis, Maria Lomeli, Lucas Hosseini, Fabio Petroni,
  Timo Schick, Jane Dwivedi-Yu, Armand Joulin, Sebastian Riedel, and Edouard
  Grave.
\newblock Few-shot learning with retrieval augmented language models.
\newblock \emph{arXiv preprint arXiv:2208.03299}, 2022.

\bibitem[Jang et~al.(2023{\natexlab{a}})Jang, Kim, Ye, Kim, Logeswaran, Lee,
  Lee, and Seo]{jang2023exploring}
Joel Jang, Seungone Kim, Seonghyeon Ye, Doyoung Kim, Lajanugen Logeswaran,
  Moontae Lee, Kyungjae Lee, and Minjoon Seo.
\newblock Exploring the benefits of training expert language models over
  instruction tuning.
\newblock In \emph{Proceedings of the International Conference of Machine
  Learning}, 2023{\natexlab{a}}.

\bibitem[Jang et~al.(2023{\natexlab{b}})Jang, Yoon, Yang, Cha, Lee, Logeswaran,
  and Seo]{jang2023knowledge}
Joel Jang, Dongkeun Yoon, Sohee Yang, Sungmin Cha, Moontae Lee, Lajanugen
  Logeswaran, and Minjoon Seo.
\newblock Knowledge unlearning for mitigating privacy risks in language models.
\newblock \emph{Proceedings of the Association for Computational Linguistics},
  2023{\natexlab{b}}.

\bibitem[{J.L. et al. v. Alphabet Inc.}(2023)]{jlversusalphabet}
{J.L. et al. v. Alphabet Inc.}
\newblock Case 3:23-cv-03416, {N.D. Cal.}, 2023.
\newblock URL
  \url{https://storage.courtlistener.com/recap/gov.uscourts.cand.415223/gov.uscourts.cand.415223.1.0.pdf}.

\bibitem[Johnson et~al.(2016)Johnson, Pollard, Shen, Lehman, Feng, Ghassemi,
  Moody, Szolovits, Anthony~Celi, and Mark]{johnson2016mimic}
Alistair~EW Johnson, Tom~J Pollard, Lu~Shen, Li-wei~H Lehman, Mengling Feng,
  Mohammad Ghassemi, Benjamin Moody, Peter Szolovits, Leo Anthony~Celi, and
  Roger~G Mark.
\newblock {MIMIC-III}, a freely accessible critical care database.
\newblock \emph{Scientific data}, 2016.

\bibitem[Johnson et~al.(2019)Johnson, Douze, and J{\'e}gou]{johnson2019billion}
Jeff Johnson, Matthijs Douze, and Herv{\'e} J{\'e}gou.
\newblock Billion-scale similarity search with {GPUs}.
\newblock \emph{IEEE Transactions on Big Data}, 2019.

\bibitem[Kandpal et~al.(2022)Kandpal, Wallace, and
  Raffel]{kandpal2022deduplicating}
Nikhil Kandpal, Eric Wallace, and Colin Raffel.
\newblock Deduplicating training data mitigates privacy risks in language
  models.
\newblock In \emph{Proceedings of the International Conference of Machine
  Learning}, 2022.

\bibitem[Khandelwal et~al.(2020)Khandelwal, Levy, Jurafsky, Zettlemoyer, and
  Lewis]{khandelwal2020generalization}
Urvashi Khandelwal, Omer Levy, Dan Jurafsky, Luke Zettlemoyer, and Mike Lewis.
\newblock Generalization through memorization: Nearest neighbor language
  models.
\newblock In \emph{Proceedings of the International Conference on Learning
  Representations}, 2020.

\bibitem[Klimt \& Yang(2004)Klimt and Yang]{klimt2004enron}
Bryan Klimt and Yiming Yang.
\newblock The {Enron} corpus: A new dataset for email classification research.
\newblock In \emph{Proceedings of European Conference of Machine Learning},
  2004.

\bibitem[Kocetkov et~al.(2023)Kocetkov, Li, allal, LI, Mou, Jernite, Mitchell,
  Ferrandis, Hughes, Wolf, Bahdanau, Werra, and de~Vries]{kocetkov2023the}
Denis Kocetkov, Raymond Li, Loubna~Ben allal, Jia LI, Chenghao Mou, Yacine
  Jernite, Margaret Mitchell, Carlos~Mu{\~n}oz Ferrandis, Sean Hughes, Thomas
  Wolf, Dzmitry Bahdanau, Leandro~Von Werra, and Harm de~Vries.
\newblock The stack: 3{TB} of permissively licensed source code.
\newblock \emph{Transactions on Machine Learning Research}, 2023.

\bibitem[Koh \& Liang(2017)Koh and Liang]{koh2017understanding}
Pang~Wei Koh and Percy Liang.
\newblock Understanding black-box predictions via influence functions.
\newblock In \emph{Proceedings of the International Conference of Machine
  Learning}, 2017.

\bibitem[Lan et~al.(2023)Lan, Cai, Wang, Huang, and Mao]{lan2023copy}
Tian Lan, Deng Cai, Yan Wang, Heyan Huang, and Xian-Ling Mao.
\newblock Copy is all you need.
\newblock In \emph{Proceedings of the International Conference on Learning
  Representations}, 2023.

\bibitem[Lee et~al.(2022)Lee, Ippolito, Nystrom, Zhang, Eck, Callison-Burch,
  and Carlini]{lee2021deduplicating}
Katherine Lee, Daphne Ippolito, Andrew Nystrom, Chiyuan Zhang, Douglas Eck,
  Chris Callison-Burch, and Nicholas Carlini.
\newblock Deduplicating training data makes language models better.
\newblock In \emph{Proceedings of the Association for Computational
  Linguistics}, 2022.

\bibitem[Lemley \& Casey(2020)Lemley and Casey]{lemley2020fair}
Mark~A Lemley and Bryan Casey.
\newblock Fair learning.
\newblock \emph{Texas Law Review}, 2020.

\bibitem[Li et~al.(2022)Li, Gururangan, Dettmers, Lewis, Althoff, Smith, and
  Zettlemoyer]{li2022branch}
Margaret Li, Suchin Gururangan, Tim Dettmers, Mike Lewis, Tim Althoff, Noah~A
  Smith, and Luke Zettlemoyer.
\newblock Branch-train-merge: Embarrassingly parallel training of expert
  language models.
\newblock \emph{arXiv preprint arXiv:2208.03306}, 2022.

\bibitem[Lin et~al.(2021)Lin, Ma, Lin, Yang, Pradeep, and
  Nogueira]{lin2021pyserini}
Jimmy Lin, Xueguang Ma, Sheng-Chieh Lin, Jheng-Hong Yang, Ronak Pradeep, and
  Rodrigo Nogueira.
\newblock Pyserini: A python toolkit for reproducible information retrieval
  research with sparse and dense representations.
\newblock In \emph{Proceedings of the ACM SIGIR Conference on Research and
  Development in Information Retrieval}, 2021.

\bibitem[Lo et~al.(2020)Lo, Wang, Neumann, Kinney, and
  Weld]{lo-etal-2020-s2orc}
Kyle Lo, Lucy~Lu Wang, Mark Neumann, Rodney Kinney, and Daniel Weld.
\newblock {S}2{ORC}: The semantic scholar open research corpus.
\newblock In \emph{Proceedings of the Association for Computational
  Linguistics}, 2020.

\bibitem[Maas et~al.(2011)Maas, Daly, Pham, Huang, Ng, and
  Potts]{maas-EtAl:2011:ACL-HLT2011}
Andrew~L. Maas, Raymond~E. Daly, Peter~T. Pham, Dan Huang, Andrew~Y. Ng, and
  Christopher Potts.
\newblock Learning word vectors for sentiment analysis.
\newblock In \emph{Proceedings of the 49th Annual Meeting of the Association
  for Computational Linguistics: Human Language Technologies}, pp.\  142--150,
  Portland, Oregon, USA, June 2011. Association for Computational Linguistics.
\newblock URL \url{http://www.aclweb.org/anthology/P11-1015}.

\bibitem[Mackenzie et~al.(2020)Mackenzie, Benham, Petri, Trippas, Culpepper,
  and Moffat]{mackenzie2020cc}
Joel Mackenzie, Rodger Benham, Matthias Petri, Johanne~R Trippas, J~Shane
  Culpepper, and Alistair Moffat.
\newblock {CC-News-En}: A large english news corpus.
\newblock In \emph{Proceedings of the ACM International Conference on
  Information and Knowledge Management}, 2020.

\bibitem[McAuley \& Leskovec(2013)McAuley and Leskovec]{mcauley2013hidden}
Julian McAuley and Jure Leskovec.
\newblock Hidden factors and hidden topics: understanding rating dimensions
  with review text.
\newblock In \emph{Proceedings of the ACM conference on Recommender systems},
  2013.

\bibitem[Metz(2022)]{metz2022lawsuit}
Cade Metz.
\newblock Lawsuit takes aim at the way {A.I.} is built.
\newblock \emph{New York Times}, 2022.

\bibitem[Min et~al.(2023)Min, Shi, Lewis, Chen, Yih, Hajishirzi, and
  Zettlemoyer]{min2022nonparametric}
Sewon Min, Weijia Shi, Mike Lewis, Xilun Chen, Wen-tau Yih, Hannaneh
  Hajishirzi, and Luke Zettlemoyer.
\newblock Nonparametric masked language modeling.
\newblock In \emph{Findings of ACL}, 2023.

\bibitem[Muennighoff et~al.(2023)Muennighoff, Rush, Barak, Scao, Piktus, Tazi,
  Pyysalo, Wolf, and Raffel]{muennighoff2023scaling}
Niklas Muennighoff, Alexander~M. Rush, Boaz Barak, Teven~Le Scao, Aleksandra
  Piktus, Nouamane Tazi, Sampo Pyysalo, Thomas Wolf, and Colin Raffel.
\newblock Scaling data-constrained language models.
\newblock \emph{arXiv preprint arXiv:2305.16264}, 2023.

\bibitem[Palen-Michel et~al.(2022)Palen-Michel, Kim, and
  Lignos]{palenmichel2022multilingual}
Chester Palen-Michel, June Kim, and Constantine Lignos.
\newblock Multilingual open text release 1: Public domain news in 44 languages.
\newblock In \emph{Proceedings of the Language Resources and Evaluation
  Conference}, 2022.

\bibitem[Pang \& Lee(2004)Pang and Lee]{pang2004sentimental}
Bo~Pang and Lillian Lee.
\newblock A sentimental education: Sentiment analysis using subjectivity
  summarization based on minimum cuts.
\newblock In \emph{Proceedings of the Association for Computational
  Linguistics}, 2004.

\bibitem[Pfeiffer et~al.(2020)Pfeiffer, Vulić, Gurevych, and
  Ruder]{pfeiffer2020madx}
Jonas Pfeiffer, Ivan Vulić, Iryna Gurevych, and Sebastian Ruder.
\newblock Mad-x: An adapter-based framework for multi-task cross-lingual
  transfer, 2020.

\bibitem[Pfeiffer et~al.(2022)Pfeiffer, Goyal, Lin, Li, Cross, Riedel, and
  Artetxe]{pfeiffer2022lifting}
Jonas Pfeiffer, Naman Goyal, Xi~Lin, Xian Li, James Cross, Sebastian Riedel,
  and Mikel Artetxe.
\newblock Lifting the curse of multilinguality by pre-training modular
  transformers.
\newblock In \emph{Conference of the North American Chapter of the Association
  for Computational Linguistics}, 2022.

\bibitem[Presser(2020)]{presser2020books}
Shawn Presser.
\newblock Books3 corpus, 2020.
\newblock URL \url{https://twitter.com/theshawwn/status/1320282149329784833}.

\bibitem[{Project Gutenberg}()]{gutenberg}
{Project Gutenberg}.
\newblock Project gutenberg.
\newblock URL \url{www.gutenberg.org}.

\bibitem[Radford et~al.(2019)Radford, Wu, Child, Luan, Amodei, Sutskever,
  et~al.]{radford2019language}
Alec Radford, Jeffrey Wu, Rewon Child, David Luan, Dario Amodei, Ilya
  Sutskever, et~al.
\newblock Language models are unsupervised multitask learners.
\newblock \emph{OpenAI blog}, 2019.

\bibitem[Raffel et~al.(2020)Raffel, Shazeer, Roberts, Lee, Narang, Matena,
  Zhou, Li, and Liu]{raffel2020exploring}
Colin Raffel, Noam Shazeer, Adam Roberts, Katherine Lee, Sharan Narang, Michael
  Matena, Yanqi Zhou, Wei Li, and Peter~J Liu.
\newblock Exploring the limits of transfer learning with a unified text-to-text
  transformer.
\newblock \emph{JMLR}, 2020.

\bibitem[Ram et~al.(2023)Ram, Levine, Dalmedigos, Muhlgay, Shashua,
  Leyton-Brown, and Shoham]{ram2023context}
Ori Ram, Yoav Levine, Itay Dalmedigos, Dor Muhlgay, Amnon Shashua, Kevin
  Leyton-Brown, and Yoav Shoham.
\newblock In-context retrieval-augmented language models.
\newblock \emph{Transactions of the Association for Computational Linguistics},
  2023.

\bibitem[Rombach et~al.(2022)Rombach, Blattmann, Lorenz, Esser, and
  Ommer]{rombach2022high}
Robin Rombach, Andreas Blattmann, Dominik Lorenz, Patrick Esser, and Bj{\"o}rn
  Ommer.
\newblock High-resolution image synthesis with latent diffusion models.
\newblock In \emph{Computer Vision and Pattern Recognition}, 2022.

\bibitem[Saxton et~al.(2019)Saxton, Grefenstette, Hill, and
  Kohli]{saxton2019analysing}
David Saxton, Edward Grefenstette, Felix Hill, and Pushmeet Kohli.
\newblock Analysing mathematical reasoning abilities of neural models.
\newblock In \emph{Proceedings of the International Conference on Learning
  Representations}, 2019.

\bibitem[Shi et~al.(2022)Shi, Michael, Gururangan, and
  Zettlemoyer]{shi2022nearest}
Weijia Shi, Julian Michael, Suchin Gururangan, and Luke Zettlemoyer.
\newblock Nearest neighbor zero-shot inference.
\newblock In \emph{Proceedings of Empirical Methods in Natural Language
  Processing}, 2022.

\bibitem[Shi et~al.(2023)Shi, Min, Yasunaga, Seo, James, Lewis, Zettlemoyer,
  and Yih]{shi2023replug}
Weijia Shi, Sewon Min, Michihiro Yasunaga, Minjoon Seo, Rich James, Mike Lewis,
  Luke Zettlemoyer, and Wen-tau Yih.
\newblock {REPLUG}: Retrieval-augmented black-box language models.
\newblock \emph{arXiv preprint arXiv:2301.12652}, 2023.

\bibitem[{Silverman et al. v. Meta Platforms, Inc.}(2023)]{silvermanmeta}
{Silverman et al. v. Meta Platforms, Inc.}
\newblock Case 3:23-cv-03417, {N.D. Cal.}, 2023.
\newblock URL
  \url{https://storage.courtlistener.com/recap/gov.uscourts.cand.415175/gov.uscourts.cand.415175.1.0.pdf}.

\bibitem[{Silverman et al. v. OpenAI, Inc.}(2023)]{silverman}
{Silverman et al. v. OpenAI, Inc.}
\newblock Case 3:23-cv-03417, {N.D. Cal.}, 2023.
\newblock URL
  \url{https://storage.courtlistener.com/recap/gov.uscourts.cand.415174/gov.uscourts.cand.415174.1.0_1.pdf}.

\bibitem[Socher et~al.(2013)Socher, Perelygin, Wu, Chuang, Manning, Ng, and
  Potts]{socher2013recursive}
Richard Socher, Alex Perelygin, Jean Wu, Jason Chuang, Christopher~D Manning,
  Andrew~Y Ng, and Christopher Potts.
\newblock Recursive deep models for semantic compositionality over a sentiment
  treebank.
\newblock In \emph{Proceedings of Empirical Methods in Natural Language
  Processing}, 2013.

\bibitem[Soldaini \& Lo(2023)Soldaini and Lo]{peS2o}
Luca Soldaini and Kyle Lo.
\newblock {peS2o (Pretraining Efficiently on S2ORC) Dataset}.
\newblock Technical report, {Allen Institute for AI}, 2023.
\newblock ODC-By, \url{https://github.com/allenai/pes2o}.

\bibitem[Together(2023)]{together2023redpajama}
Together.
\newblock {RedPajama}: An open source recipe to reproduce {LLaMA} training
  dataset, 2023.
\newblock URL \url{https://github.com/togethercomputer/RedPajama-Data}.

\bibitem[Touvron et~al.(2023)Touvron, Lavril, Izacard, Martinet, Lachaux,
  Lacroix, Rozière, Goyal, Hambro, Azhar, Rodriguez, Joulin, Grave, and
  Lample]{touvron2023llama}
Hugo Touvron, Thibaut Lavril, Gautier Izacard, Xavier Martinet, Marie-Anne
  Lachaux, Timothée Lacroix, Baptiste Rozière, Naman Goyal, Eric Hambro,
  Faisal Azhar, Aurelien Rodriguez, Armand Joulin, Edouard Grave, and Guillaume
  Lample.
\newblock {LLaMA}: Open and efficient foundation language models.
\newblock \emph{arXiv preprint arXiv:2302.13971}, 2023.

\bibitem[Tremblay et al.~v. OpenAI(2023)]{tremblay}
Inc. Tremblay et al.~v. OpenAI.
\newblock Case 3:23-cv-03223 , {N.D. Cal.}, 2023.
\newblock URL
  \url{https://storage.courtlistener.com/recap/gov.uscourts.cand.414822/gov.uscourts.cand.414822.1.0.pdf}.

\bibitem[Vaswani et~al.(2017)Vaswani, Shazeer, Parmar, Uszkoreit, Jones, Gomez,
  Kaiser, and Polosukhin]{vaswani2017attention}
Ashish Vaswani, Noam Shazeer, Niki Parmar, Jakob Uszkoreit, Llion Jones,
  Aidan~N. Gomez, Lukasz Kaiser, and Illia Polosukhin.
\newblock Attention is all you need.
\newblock In \emph{Proceedings of Advances in Neural Information Processing
  Systems}, 2017.

\bibitem[Vincent(2023)]{vincent2023}
James Vincent.
\newblock Getty images sues {AI} art generator stable diffusion in the us for
  copyright infringement, 2023.
\newblock URL
  \url{https://www.theverge.com/2023/2/6/23587393/ai-art-copyright-lawsuit-getty-images-stable-diffusion}.

\bibitem[Vyas et~al.(2023)Vyas, Kakade, and Barak]{vyas2023provable}
Nikhil Vyas, Sham Kakade, and Boaz Barak.
\newblock Provable copyright protection for generative models.
\newblock 2023.

\bibitem[Xie et~al.(2023)Xie, Pham, Dong, Du, Liu, Lu, Liang, Le, Ma, and
  Yu]{xie2023doremi}
Sang~Michael Xie, Hieu Pham, Xuanyi Dong, Nan Du, Hanxiao Liu, Yifeng Lu, Percy
  Liang, Quoc~V Le, Tengyu Ma, and Adams~Wei Yu.
\newblock {DoReMi}: Optimizing data mixtures speeds up language model
  pretraining.
\newblock \emph{arXiv preprint arXiv:2305.10429}, 2023.

\bibitem[Yu et~al.(2023)Yu, Shi, Pasunuru, and Miller]{meta2023}
Lili Yu, Bowen Shi, Ram Pasunuru, and Benjamin Miller.
\newblock {Scaling Autoregressive Multi-Modal Models: Pretraining and
  Instruction Tuning}, 2023.

\bibitem[Zhang et~al.(2021)Zhang, Ippolito, Lee, Jagielski, Tram{\`e}r, and
  Carlini]{zhang2021counterfactual}
Chiyuan Zhang, Daphne Ippolito, Katherine Lee, Matthew Jagielski, Florian
  Tram{\`e}r, and Nicholas Carlini.
\newblock Counterfactual memorization in neural language models.
\newblock \emph{arXiv preprint arXiv:2112.12938}, 2021.

\bibitem[Zhang et~al.(2023)Zhang, Finckenberg-Broman, Hoang, Pan, Xing,
  Staples, and Xu]{zhang2023right}
Dawen Zhang, Pamela Finckenberg-Broman, Thong Hoang, Shidong Pan, Zhenchang
  Xing, Mark Staples, and Xiwei Xu.
\newblock Right to be forgotten in the era of large language models:
  Implications, challenges, and solutions.
\newblock \emph{arXiv preprint arXiv:2307.03941}, 2023.

\bibitem[Zhang et~al.(2015)Zhang, Zhao, and LeCun]{zhang2015character}
Xiang Zhang, Junbo Zhao, and Yann LeCun.
\newblock Character-level convolutional networks for text classification.
\newblock In \emph{Proceedings of Advances in Neural Information Processing
  Systems}, 2015.

\bibitem[Zhong et~al.(2022)Zhong, Lei, and Chen]{zhong2022training}
Zexuan Zhong, Tao Lei, and Danqi Chen.
\newblock Training language models with memory augmentation.
\newblock In \emph{Proceedings of Empirical Methods in Natural Language
  Processing}, 2022.

\bibitem[Zhu et~al.(2015)Zhu, Kiros, Zemel, Salakhutdinov, Urtasun, Torralba,
  and Fidler]{zhu2015aligning}
Yukun Zhu, Ryan Kiros, Rich Zemel, Ruslan Salakhutdinov, Raquel Urtasun,
  Antonio Torralba, and Sanja Fidler.
\newblock Aligning books and movies: Towards story-like visual explanations by
  watching movies and reading books.
\newblock In \emph{Computer Vision and Pattern Recognition}, 2015.

\end{thebibliography}
\bibliographystyle{iclr2024_conference}

\clearpage
\appendix

\section{Model details}\label{app:model-details}

\paragraph{Details on the parametric component \modelname.}

Table~\ref{tab:model_hyperparams} reports the hyperparameters for the parametric component of \modelname. We keep these hyperparameters fixed for all parametric models that we report in this paper. We follow the model architecture of LLaMa \citep{touvron2023llama}, and we use the GPT-NeoX-20B tokenizer \citep{black2022gptneox20b}, with 50432 BPE types. During training, we use 2,048 token sequences that are packed across document boundaries, and we pre-pend a beginning-of-text token to every document. We use weight decay of 0.1, the Adam optimizer with $\beta_2=$ 0.95, 2,000 steps of warmup, with a cosine learning rate scheduler. We train for multiple epochs in each dataset, tracking validation perplexity every 10B tokens, and perform early stopping. We train our \PD, \PD\SW\ and \PD\SW\BY~models for 60B, 250B, and 350B tokens in total, respectively. 

\begin{table*}[h]
    \centering \footnotesize
    \begin{tabular}{lrrrrr}
        \toprule
        \bf Model & \bf \#L & \bf \#H & \bf $\text{d}_{\text{model}}$ & \bf LR & \bf Batch \\
        \midrule
        1.3B & 24 & 16 & 2048 & 1e-3 & 2.6M \\
        \bottomrule
    \end{tabular}
    \caption{Basic hyperparameters for the parametric component of \modelname.}\label{tab:model_hyperparams}
\end{table*}

\paragraph{Details on the nonparametric component of \modelname.}
For \knnlm, we use IndexIVFPQ which quantizes vectors into 64-bytes and clusters them into 4,096 centroids, learned from 1 million sampled vectors, following \citet{khandelwal2020generalization}.
Instead of recomputing the exact L2 distance using the original embeddings, we use the L2 distance beteen quantized vectors returned by the FAISS index (ablations in Appendix~\ref{app:additional-nonparametric}). Since their scale is not preserved, we use $\frac{d(\mathbf{x}_\mathrm{q}, \mathbf{y}_\mathrm{q})}{\tau}$ as a proxy of $d(\mathbf{x}, \mathbf{y})$, where $\mathbf{x}_\mathrm{q}$ and $\mathbf{y}_\mathrm{q}$ are vectors quantized from $\mathbf{x}$ and $\mathbf{y}$. Hyperparameters, including $k$, $\lambda$, and $\tau$, are chosen based on the validation data in a domain-specific manner.

Table~\ref{tab:datastore-statistics} reports the datastore statistics for both \riclm\ and \knnlm, as well as hyperparameter values for \knnlm\ ($\lambda, k, \tau$).
Due to the resource constraints, the datastore size is capped to up to 10\% of the PILE training data (and to 1024.0M tokens in the case of \knnlm), but future work can investigate further scaling the datastore.

\begin{table*}[h]
    \centering \footnotesize
    \begin{tabular}{lrrrrrrrr}
        \toprule
            \multirow{2}{*}{Data} & \multicolumn{2}{c}{\riclm} 
            && \multicolumn{5}{c}{\knnlm} \\
            \cmidrule(lr){2-3} \cmidrule(lr){5-9}
            & \# tokens & \# blocks && \# tokens && $\lambda$ & $k$ & $\tau$ \\
        \midrule
            Github          & 3084.3M  & 6.0M &&1024.0M&    & 0.2 & 128 & 10.0 \\
            NIH ExPorter    & 72.2M & 0.1M && 72.2M &             &0.3&32,768&20.0\\
            Wikipedia       & 1177.9M & 2.3M &&1024.0M&     &0.3&4,096&20.0\\
            CC News         & 382.2M & 0.7M &&382.2M&       &0.7&4,096&20.0\\
            Books3          & 1424.7M & 2.8M &&1024.0M&      &0.2&4,096&25.0\\
            Enron Emails    & 45.0M & 0.1M && \underline{45.0M} &          &\underline{0.5}&\underline{4,096}&\underline{1.0}\\
            Amazon          & 1214.3M & 2.4M &&1024.0M&     &0.5&32,768&20.0\\
            MIMIC-III       & 519.5M & 1.0M && 519.5M&       &0.7&1,024&15.0\\
        \bottomrule
    \end{tabular}
    \caption{Datastore statistics as well as hyperparameter values for \knnlm.
    \underline{Underline} indicates exact nearest neighbor search (instead of approximate) was performed for \knnlm\ because the datastore is small enough.
    Hyperparameters are chosen based on the validation data of each domain.
    }\label{tab:datastore-statistics}
\end{table*}

\begin{table*}[t]
    \caption{Datastore and hyperparameter values for \knnlm\ evaluated on downstream tasks.
    Hyperparameters (the choice of datastore, $\lambda$, $k$ and $\tau$) are chosen based on the validation data.
    }\label{tab:knn-prompt-statistics} \vspace{-.3em}
    \centering \myfontsize 
\begin{tabular*}{0.99\linewidth}{lcccc|c|c}
\toprule
Task & Datastore & $\lambda$ & $k$ & $\tau$ & Template & Label Words \\
\midrule
AGN & Wikipedia & 0.8 & 1024 & 3 & The topic of the text is & politics, sports, business, technology \\
Dbpedia & Amazon & 0.9 & 4096 & 3 & The topic of the text is & company, school, artist, athlete ... (14 in total) \\
SST-2 & IMDB & 0.9 & 8192 & 3 & It was  & great, terrible \\
MR & IMDB & 0.9 & 8192 & 3 & It was  & great, terrible \\
RT & IMDB & 0.9 & 8192 & 3 & It was  & great, terrible \\
CR & Amazon & 0.7 & 1024 & 1 & It was  & great, terrible \\
Yelp & IMDB & 0.9 & 8192 & 3 & It was  & great, terrible \\
Amz & IMDB & 0.9 & 8192 & 3 & It was  & great, terrible \\
RTE & Amazon & 0.9 & 1024 & 3 & true or false? answer: & true, false \\
HYP & CC News & 0.1 & 1024 & 1 & neutral or partisan? answer:? & neutral, partisan \\

\bottomrule
\end{tabular*}
\end{table*}

\paragraph{Details of Downstream Task Evaluation.}
We perform zero-shot prompting for nine text classification datasets: AGNews~\citep{zhang2015character}, Yahoo~\citep{zhang2015character}, Subj~\citep{pang2004sentimental}, SST-2~\citep{socher2013recursive}, MR~\citep{pang2004sentimental}, Rotten Tomatoes (RT), CR~\citep{hu2004mining}, Amazon polarity (Amz, \citet{mcauley2013hidden}) and RTE~\citep{dagan2005pascal}.
The tasks range from topic classification and sentiment analysis to subjectivity classification and textual entailment.

We use the templates and label words to map the task into a sentence completion problem, following the standard from literature~\citep{radford2019language,gpt3,holtzman-etal-2021-surface}.
These templates and label words are taken from \citet{shi2022nearest}, reported in Table~\ref{tab:knn-prompt-statistics}.
For both parametric LMs and $k$NN-LM, we apply the domain-conditional PMI scoring~\citep{holtzman-etal-2021-surface} for determining the probability of each label.
For $k$NN-LM, we follow a method from~\citet{shi2022nearest} which employs the fuzzy verbalizers to expand the token set associated with each output label in our task.
Also following \citet{shi2022nearest}, we use IMDB~\citep{maas-EtAl:2011:ACL-HLT2011} with 8 million tokens as an additional datastore.
We perform hyperparameter search on the validation dataset of each task, considering $k \in \{128, 512, 4196, 8192\}$, $\tau \in \{1, 3, 5, 10, 40, 80\}$, and different choice of datastores. 
The chosen hyperparameters are reported in Table~\ref{tab:knn-prompt-statistics}.

\section{Additional Data Analysis}\label{app:additional-data-analysis}

\begin{figure}
    \centering
    \includegraphics[scale=0.8]{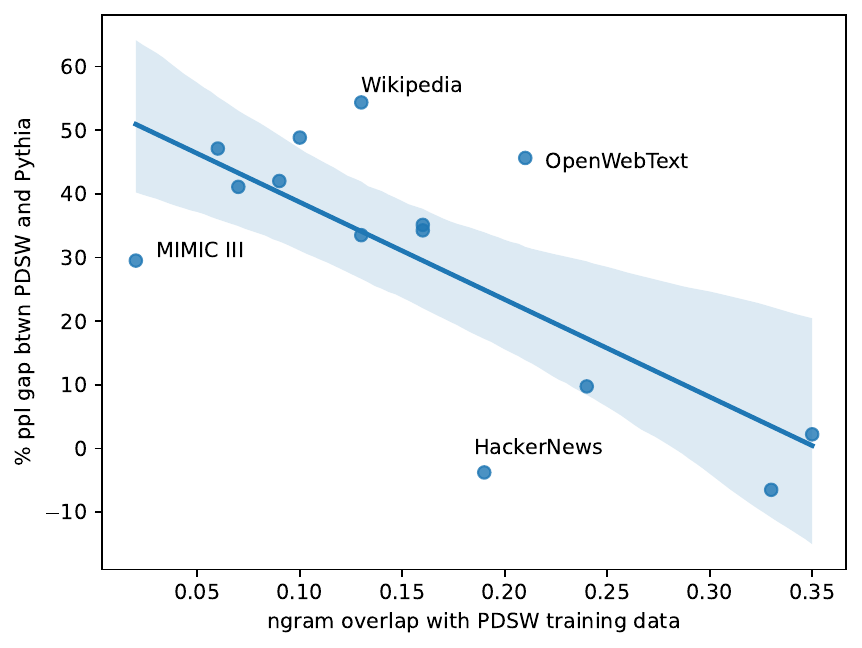}
    \caption{There is a strong negative correlation between ngram overlap of a domain with the \PD\SW~training data and the perplexity gap between the \PD\SW~LM and Pythia ($r$=-0.72, $p<$ 0.005).}
    \label{fig:ngram_overlap}
\end{figure}

\tightparagraph{N-gram overlap.}
Table~\ref{tab:ngram-overlap-ours-pile} displays the full matrix of unigram and bi-gram overlap between the \dataname~training domains and the Pile validation domains. We sample up to 10M tokens in each data source, remove stopwords, and only consider unigrams and bigrams that appear in at least three documents. We also show a scatterplot that describes the relationship between ngram overlap and the performance gap between \PD\SW~and Pythia in Figure \ref{fig:ngram_overlap}.


\section{Additional Experimental Results}\label{app:additional-results}

\begin{table}[t]
    \myfontsize 
    \centering
    \begin{tabular}{l R{1.4cm} R{1.4cm} R{1.4cm} R{1.4cm}}
        \toprule
            Eval data & \PD & \PD\SW & \PD\SW\BY & Pythia \\
        \midrule
            FreeLaw         & \ID{5.3} & \ID{5.7} & \ID{6.5} & \ID{5.6} \\
            Gutenberg       & \ID{14.6} & \ID{11.9} & \ID{13.4} & \ID{12.7} \\
            HackerNews      & \OD{36.6} & \ID{12.1} & \ID{13.2} & \ID{12.5} \\
            Github          & \OD{13.3} & \SD{2.6} & \SD{2.7} & \ID{2.4} \\
            NIH ExPorter    & \OD{28.6} & \SD{19.3} & \SD{15.1} & \ID{11.2} \\
            PhilPapers      & \OD{55.2} & \SD{24.2} & \SD{16.5} & \ID{14.3} \\
            Wikipedia       & \OD{27.9} & \OD{19.7} & \ID{11.1} & \ID{9.0} \\
            CC News         & \OD{30.8} & \OD{21.3} & \OD{19.3} & \ID{10.9} \\
            BookCorpus2     & \OD{25.2} & \OD{19.2} & \OD{20.2} & \ID{12.8} \\
            Books3          & \OD{25.9} & \OD{18.7} & \OD{18.1} & \ID{12.4} \\
            OpenWebText2    & \OD{38.1} & \OD{21.2} & \OD{18.8} & \ID{11.5} \\
            Enron Emails    & \OD{19.9} & \OD{14.3} & \OD{14.5} & \ID{7.6} \\
            Amazon          & \OD{81.9} & \OD{34.7} & \OD{37.0} & \ID{22.8} \\
            MIMIC-III       & \OD{18.2} & \OD{16.4} & \OD{13.6} & \OD{11.5} \\
        \midrule
            Average         & \num{30.1} & \num{17.2} & \num{15.7} & \num{11.2} \\
        \bottomrule
    \end{tabular}
    \caption{
        Perplexity on the parametric LMs trained on \PD{}, \PD{}\SW{}, and \PD{}\SW{}\BY{}, as well as Pythia 1.4B, a model trained with similar amounts of compute but on non-permissive data.
        We use
        \textcolor{bluex!25}{$\blacksquare$},
        \textcolor{orange!25}{$\blacksquare$},
        and \textcolor{yellow!25}{$\blacksquare$} to indicate text that is in-domain, out-of-domain, or out-of-domain but has relevant data in-domain data (e.g., non-permissive Github code versus our permissive training code). 
        Reported on the validation data; see Table~\ref{tab:results-parameteric} for results on the test data.
    }\label{tab:results-parameteric-val}
\end{table}

\begin{table}[t]
    \myfontsize 
    \centering
    \begin{tabular}{l R{1.4cm} R{1.4cm} R{1.4cm} R{1.4cm} } 
        \toprule
            \multirow{2}{*}{Eval data} & \multicolumn{3}{c}{\PD\SW} & \multicolumn{1}{c}{Pythia} \\
            \cmidrule(lr){2-4} \cmidrule(lr){5-5}
            & Prm-only & \knnlm & \riclm & Prm-only  \\
        \midrule
            Github          & \SD{2.6}  & \SD{2.4} & \SD{2.4} & \ID{2.4}  \\
            NIH ExPorter    & \SD{19.3} & \SD{14.9} & \SD{18.5} & \ID{11.2} \\
            Wikipedia       & \OD{19.7} & \OD{14.1} & \OD{18.9} & \ID{9.0} \\
            CC News         & \OD{21.3} & \OD{7.1} & \OD{14.8} & \ID{10.9} \\
            Books3          & \OD{18.8} & \OD{17.3} & \OD{18.5} & \ID{12.5} \\
            Enron Emails    & \OD{14.3} & \OD{6.7} & \OD{11.1} & \ID{7.6} \\
            Amazon          & \OD{34.7} & \OD{26.2} & \OD{33.7} & \ID{22.8} \\
            MIMIC-III       & \OD{16.3} & \OD{7.2} & \OD{14.1} & \OD{11.5} \\
        \midrule
            Average         & \num{18.4} & \num{12.0} & \num{16.5} & \num{11.0} \\
        \bottomrule
    \end{tabular}
    \caption{
        Perplexity of parametric LMs (Prm-only), \knnlm\ and \riclm; \textcolor{bluex!25}{$\blacksquare$}
        indicates in-domain;
        \textcolor{orange!25}{$\blacksquare$}
        indicates out-of-domain;
        \textcolor{yellow!25}{$\blacksquare$} indicates out-of-domain but has relevant data in-domain.
        Reported on the validaiton data; see Table~\ref{tab:results-nonparameteric} for results on the test data.
    }\label{tab:results-nonparameteric-val}
\end{table}

\begin{table*}
\centering \footnotesize
    \hfill
    \begin{tabular}{lrr}
        \toprule
            {\shortstack{Data}} & {\shortstack{\PD\SW \\ w/o upsampling}} & {\shortstack{\PD\SW \\ w upsampling}} \\
        \midrule
            FreeLaw         & \ID{4.9} & \ID{5.7} \\
            Github          & \SD{2.4} & \SD{2.6} \\
            NIH ExPorter    & \SD{20.0} & \SD{19.3} \\
            PhilPapers      & \SD{23.9} & \SD{24.2} \\
            Wikipedia       & \OD{19.9} & \OD{19.7} \\
            CC News         & \OD{21.8} & \OD{21.3} \\
            BookCorpus2     & \OD{19.4} & \OD{19.2} \\
            OpenWebText2    & \OD{21.0} & \OD{21.2} \\
            Enron Emails    & \OD{13.5} & \OD{14.3} \\
            Amazon          & \OD{35.7} & \OD{34.7} \\
        \bottomrule
    \end{tabular}
    \hfill
    \begin{tabular}{lrrr}
        \toprule
            {\shortstack{Data}} & {\shortstack{\PD}} &  
            {\shortstack{\PD\SW \\ w/o code}} & 
            {\shortstack{\PD\SW}} \\
        \midrule
            FreeLaw         &  \ID{5.3} & \ID{5.7} & \ID{5.7} \\
            Github          & \OD{13.3} &  \OD{8.2} & \SD{2.6} \\
            NIH ExPorter    & \OD{28.6} & \SD{26.2} & \SD{19.3} \\
            PhilPapers      & \OD{55.2} & \SD{36.4} & \SD{24.2} \\
            Wikipedia       & \OD{27.9} & \OD{26.5} & \OD{19.7}\\
            CC News         & \OD{30.8} & \OD{28.8} & \OD{21.3}\\
            BookCorpus2     & \OD{25.2} & \OD{23.8} & \OD{19.2}\\
            OpenWebText2    & \OD{38.1} 
 & \OD{31.7} & \OD{21.2}\\
            Enron Emails    & \OD{19.9} & \OD{18.5} & \OD{14.3}\\
            Amazon          & \OD{81.9} & \OD{46.1} & \OD{34.7}\\
        \bottomrule
    \end{tabular}
    \hfill
    \caption{
        \textbf{(Left)} {Effect of re-weighting rare domains,} comparing models trained on \datanamePDSW\ with and without upsampling.
        \textbf{(Right)} {Effect of \SW{} data, with and without explicit source code}---we train an LM with \SW{} data but remove all of the actual source code (i.e., we leave Hacker News, Ubuntu IRC, Deepmind Math, and AMPS).
        Both tables report perplexity on the validation data.
    }\label{tab:ablation}
\end{table*}

\paragraph{Results on the validation data.}
Table~\ref{tab:results-parameteric-val} reports perplexity of the parametric LMs on the validation data that is analogous to Table~\ref{tab:results-parameteric}.
Table~\ref{tab:results-nonparameteric-val} reports perplexity of both parametric and nonparametric LMs on the validation data that is analogous to Table~\ref{tab:results-nonparameteric}.
Findings based on the validation data and on the test data are largely consistent.

\subsection{Ablations: Parametric Component (Section~\ref{subsec:results-parametric})}\label{app:additional-parametric}

\tightparagraph{Effect of upsampling low-resource data.}
As described in \S\ref{subsec:exp-setup}, since \dataname\ has an extremely skewed distribution of domains, we upsample less-representative domains during training.
Table~\ref{tab:ablation} (left) compares the models trained on \PD\SW\ with and without domain upweighting.
In-domain datasets that are not upweighted, e.g., FreeLaw, see slight degration in performance. On out-of-doain datasets, there is no significant differences, although the model with upsampling is marginally better (19.6 vs. 19.7 when averaged over 9 out-of-domain datasets). We note that we did not tune the upweighting ratio nor explore alternative upweighting approaches~\citep{xie2023doremi} due to resource constraints, and leave them for future work.

\tightparagraph{59B tokens of source code significantly help.}
When using \SW{}, a substantial $59.1\%$ of the training data is actual source code. To determine \SW{} provides such large gains, we also run an ablation where we include \SW{} data but exclude all of the actual source code, i.e., we only include Hacker News, Ubuntu IRC, Deepmind Math, and AMPS on top of the \PD{} data. This leaves models trained on 99.6B tokens for \datanamePDSW\ and 40.7B for \datanamePDSW\ excluding source code. 
Table~\ref{tab:ablation} (right) report results on a subset of the validation domains. Including source code provide significant benefits for certain test datasets, e.g., nearly a 20 point improvement in perplexity on PhilPapers, likely because it significantly increases the size of the training data.

\begin{figure*}[t]
\centering
\includegraphics[trim={0cm, 0cm, 0cm, 0},clip,width=0.35\columnwidth]{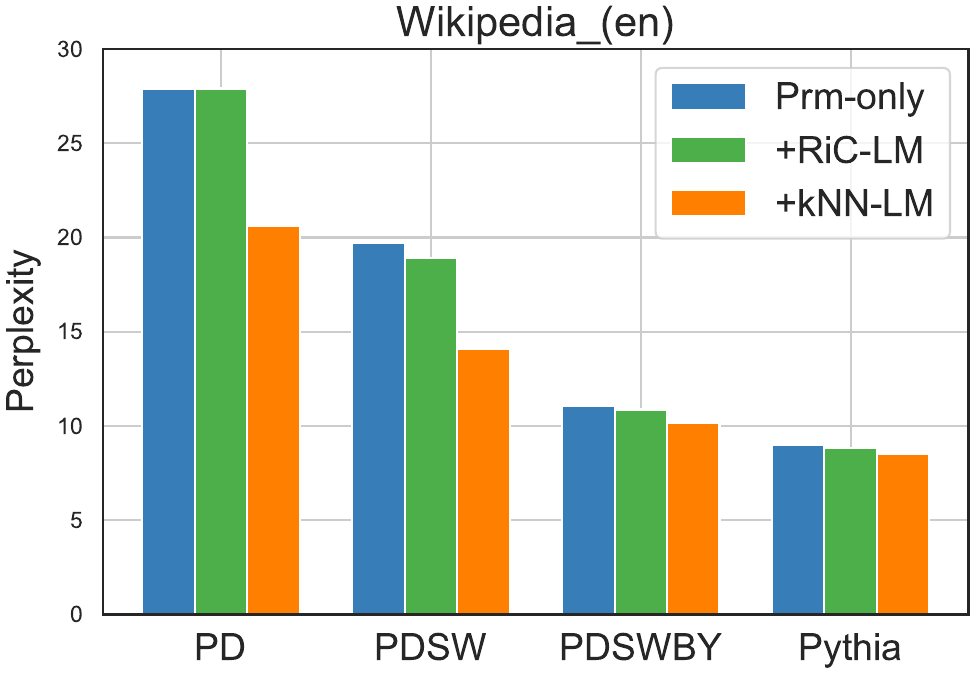}
\includegraphics[trim={1.5cm, 0cm, 0cm, 0},clip,width=0.318\columnwidth]{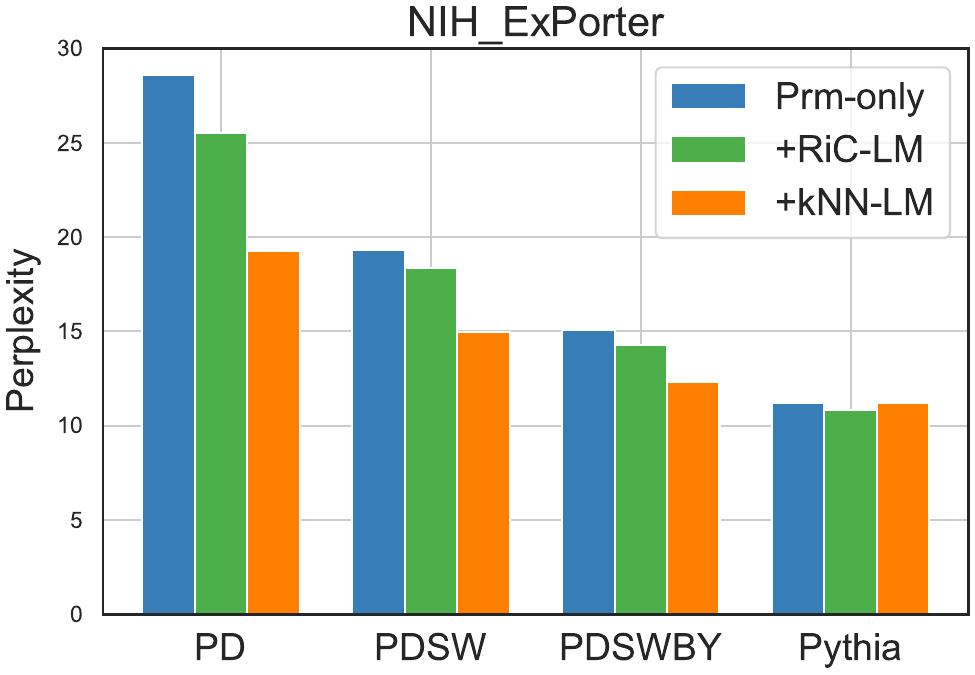}
\includegraphics[trim={1.5cm, 0cm, 0cm, 0},clip,width=0.318\columnwidth]{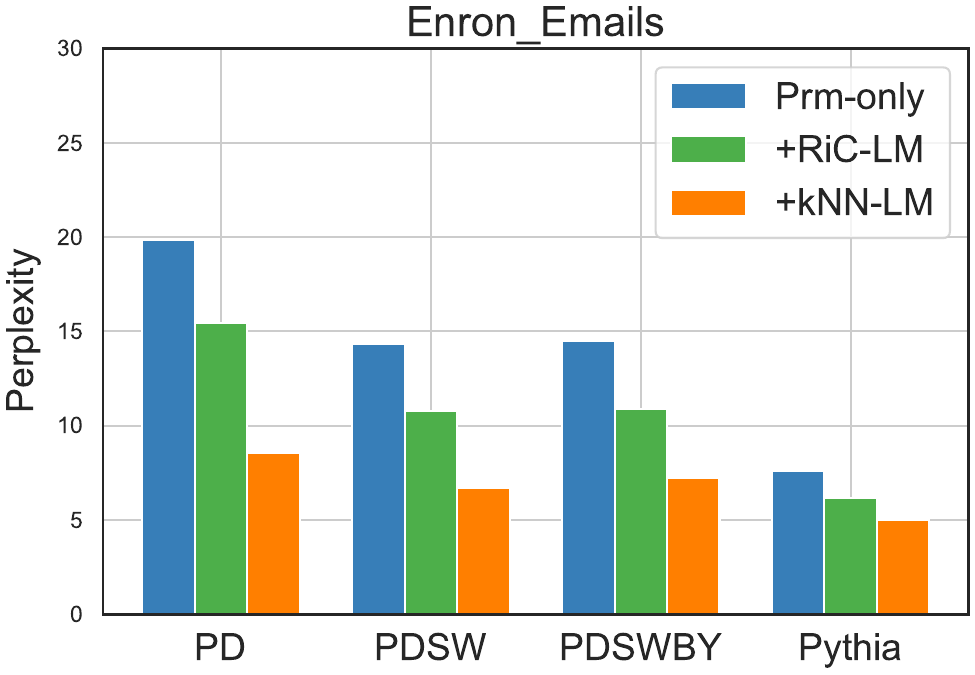}
\vspace{-1em}
\caption{Results on different variants of models: \PD, \PD\SW\ and \PD\SW\BY\ variants of \modelname\ as well as Pythia.
Adding a nonparametric component through either \riclm\ and \knnlm\ helps and \knnlm\ is overall better than \riclm, consistently across all models and evaluation datasets.
}
\label{fig:abl-lm-nonparametric}
\end{figure*}

\subsection{Ablations: Adding the Nonparametric Component (Section~\ref{subsec:results-nonparametric})}\label{app:additional-nonparametric}

\paragraph{Impact of adding a nonparametric component across varying LMs.}
We compare parametric-only LM, \riclm\ and \knnlm\ over four different LMs: \PD, \PD\SW\ and \PD\SW\BY\ variants of \modelname\ as well as Pythia.
Figure~\ref{fig:abl-lm-nonparametric} reports their results on three evaluation datasets: Wikipedia, NIH ExPorter and Enron Emails.
Findings from Section~\ref{subsec:results-nonparametric} hold across all models: both \riclm\ and \knnlm\ are consistently better than the parametric-only LM, and \knnlm\ overall achieves the best performance.
For instance, \knnlm\ allows the model to be comparable to or outperform the one-level relaxed variant, e.g., a \PD-based \knnlm\ is comparable to a \PD\SW-based parametric LM, and \PD\SW-based \knnlm\ is comparable to a \PD\SW\BY-based parametric LM.

\begin{table*}[t]
\centering \footnotesize
\setlength{\tabcolsep}{5pt}
    \begin{tabular}{lrrrrrrrrr}
        \toprule
            \multirow{2}{*}{Data} & \multicolumn{4}{c}{\PD\SW} & \multicolumn{4}{c}{Pythia} \\
            \cmidrule(lr){2-5} \cmidrule(lr){6-9}
            & basic & ensbl-10 & concat-2 & concat-next & basic & ensbl-10 & concat-2 & concat-next \\
        \midrule
           CC News & \OD{14.8} & \OD{13.5} & \OD{17.0} & \OD{18.8} & \ID{8.2} & \ID{7.9} & \ID{9.2} & \ID{9.9} \\
           Enron Emails & \OD{11.1} & \OD{10.0} & \OD{12.8} & \OD{13.4} & \ID{6.3} & \ID{6.117} & \ID{7.1} & \ID{7.3} \\
        \bottomrule
    \end{tabular}
    \caption{
        Ablations on different variants of \riclm{}s.
        Perplexity on the validation data reported.
        {\em ensbl-10} is 10x slower than other three methods.
    }\label{tab:ablation-riclm}
\end{table*}

\tightparagraph{Ablations on variants of \riclm.}
We consider four different variants of \riclm.
(1) The \textbf{basic} is the method described in \S\ref{subsec:non-parametric-lm}, which uses text blocks with a length of $L$ each. At inference, it takes the top 1 text block from the datastore and feeds it to the LM, i.e., $P_\mathrm{LM}(y|\hat{b}, x)$ where $x$ is the input and $\hat{b}$ is the top 1 text block. 
(2) The \textbf{ensbl-$\bm{k}$} ($k=10$) variants is also based on text blocks with a length of $L$ each. At inference, it takes the top $k$ text blocks from the datastore, feeds it to the LM in parallel and aggregates the probability distributions, e.g., $\frac{1}{k}\sum\limits_{1 \leq i \leq k} P_\mathrm{LM}(y|\hat{b}_i, x)$ where $\hat{b}_1...\hat{b}_k$ are the top $k$ text blocks.
This follows a method from \citet{shi2023replug}. 
(3) The \textbf{concat-$\bm{k}$} ($k=2$) variant uses text blocks with a length of $\frac{L}{k}$ each.
At inference, it takes the top $k$ text blocks from the datastore, concatenates them in a reverse order, and feeds it into the LM, e.g., $P_\mathrm{LM}(y|\hat{b}_k, \cdots, \hat{b}_1, x)$ where $\hat{b}_1...\hat{b}_k$ are the top $k$ text blocks.
(4) The \textbf{concat-next} variant uses text blocks with a length of $\frac{L}{2}$ each.
At inference, it takes the top 1 text block from the datastore, and concatenates the text block and the subsequent text block in a datastore. It then feeds it into the LM. This is based on the intuition that the continuation of the text block that is most similar to the query can be useful for the continuation of the query; \citet{borgeaud2022improving} has explored a similar approach based on the same intuition.
We use $L=1024$ for all variants. It is worth noting that
the \textbf{ensbl-$\bm{k}$} variant has run-time that is approximately $k$ times of run-time of the \textbf{basic}, \textbf{concat-$\bm{k}$} and \textbf{concat-next}.

Results are reported in Table~\ref{tab:ablation-riclm}.
The concat-2 and concat-next variants perform poorly, while the ensbl-10 outperforms the basic variant. However, we reached the conclusion that the significant run-time cost (i.e., 20x compared to a parametric LM) does not justify the improvements, and thus, we primarily use the basic variant for the remaining experiments. Future work may involve re-evaluating models using the ensbl-$k$ approach or enhancing its run-time efficiency.

\tightparagraph{Effect of scaling the datastore in-domain and out-of-domain.}
\S\ref{subsec:results-nonparametric} shows that performance of both \knnlm\ and \riclm\ rapidly improves as the datastore size grows, and \knnlm\ improves more rapidly than \riclm\ does. This evaluation is mainly done with \modelname\ where the test domains are out-of-domain. Does this trend hold when the test domains are in-domain? To answer this question, we examine effect of scaling the datastore with Pythia 1.4B, where all of our test datasets can be considered in-domain.

\begin{figure*}[t]
\centering
\includegraphics[width=0.42\columnwidth]{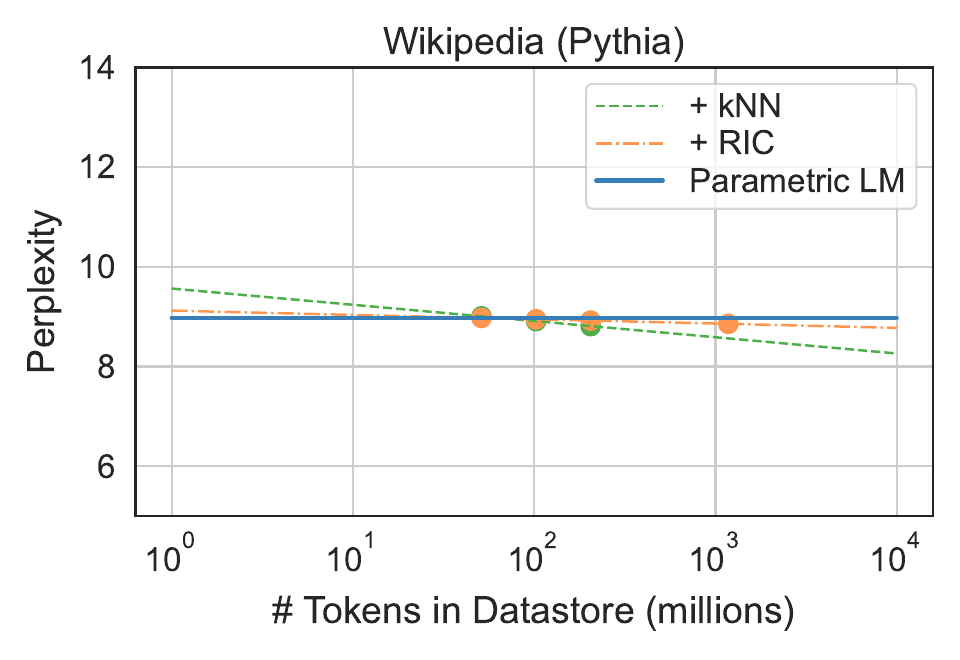}
\includegraphics[width=0.42\columnwidth]{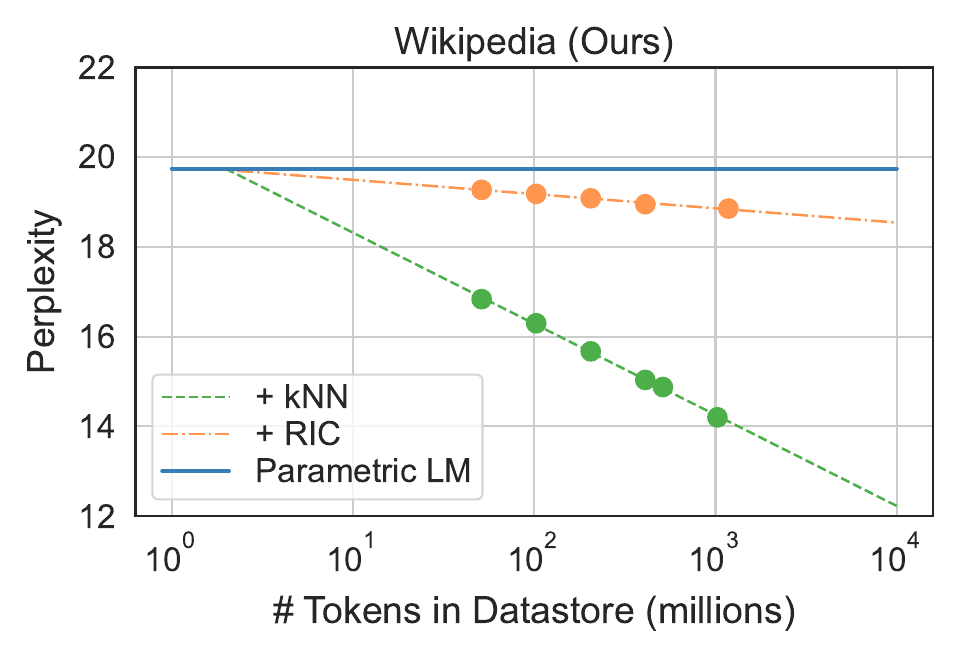}
\includegraphics[width=0.42\columnwidth]{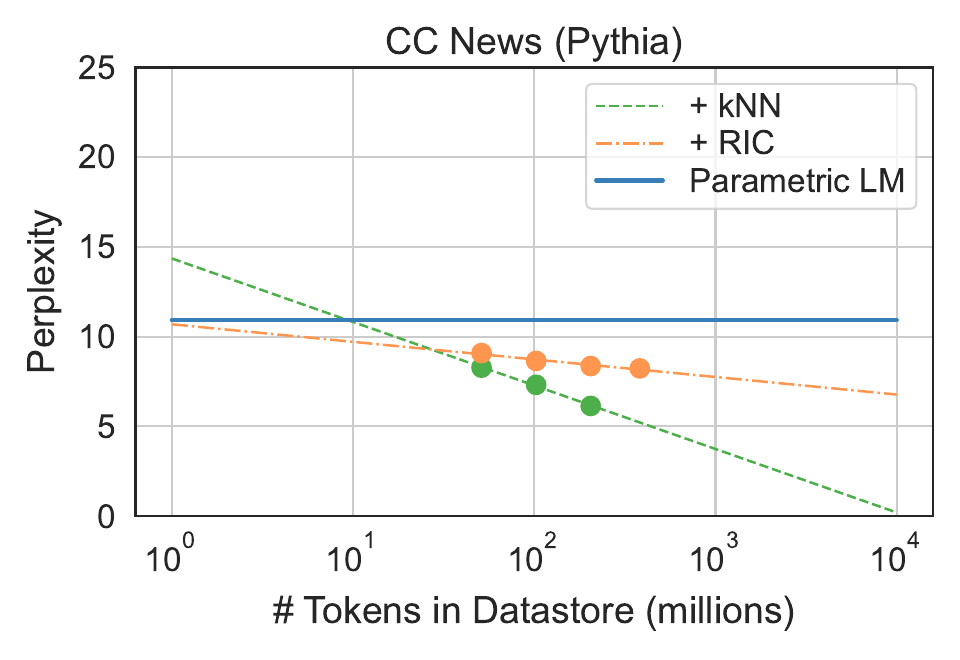}
\includegraphics[width=0.42\columnwidth]{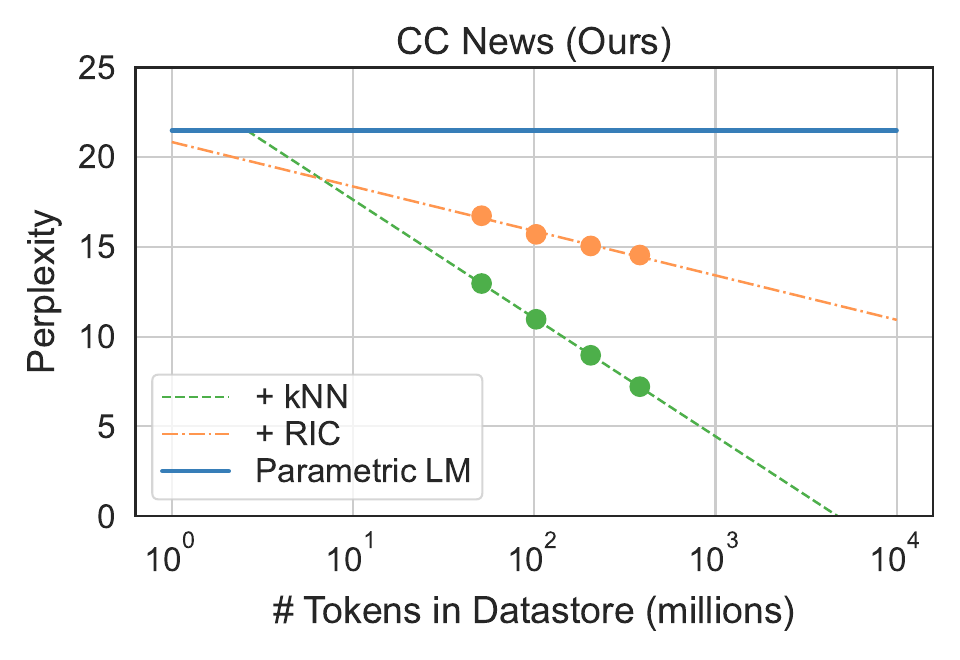}
\includegraphics[width=0.42\columnwidth]{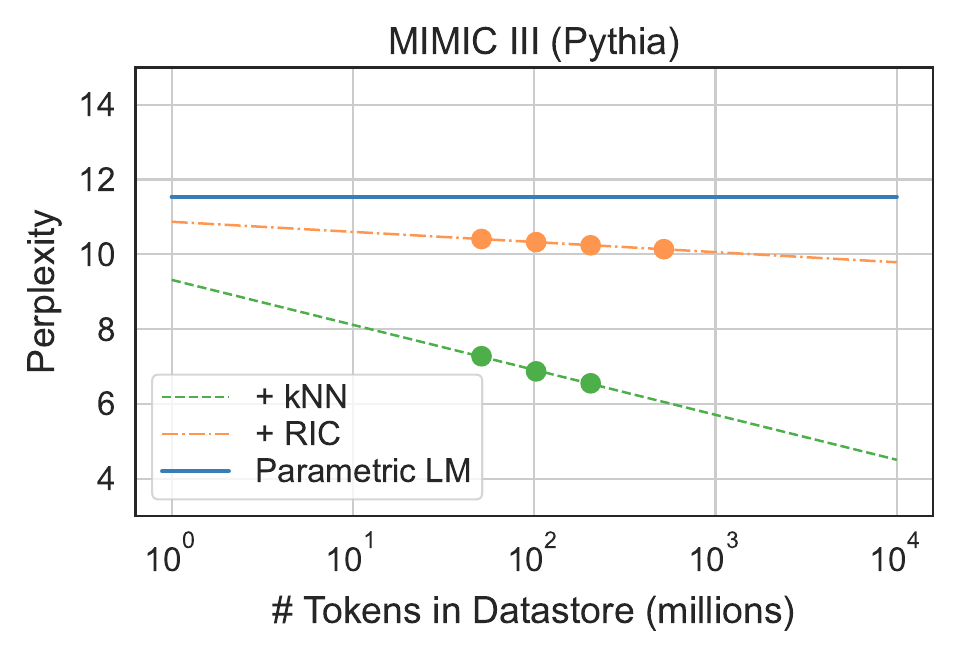}
\includegraphics[width=0.42\columnwidth]{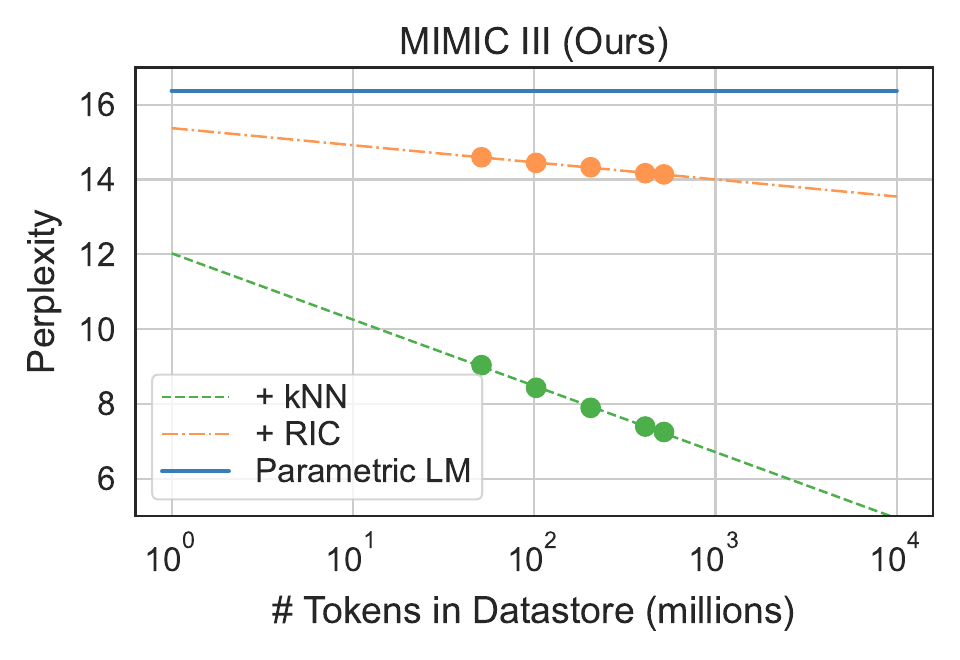}
\includegraphics[width=0.42\columnwidth]{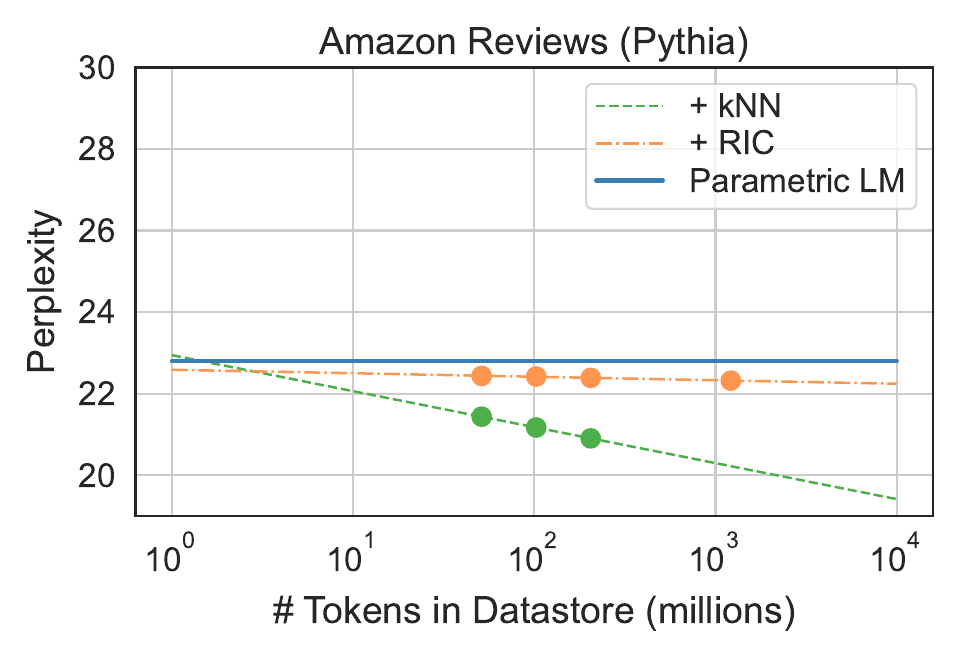}
\includegraphics[width=0.42\columnwidth]{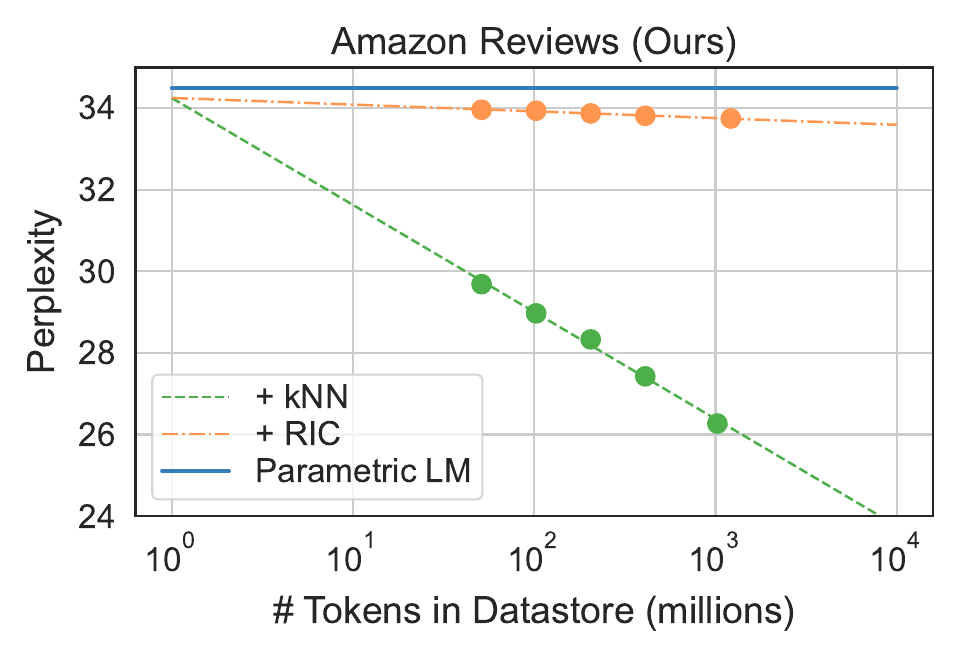}
\includegraphics[width=0.42\columnwidth]{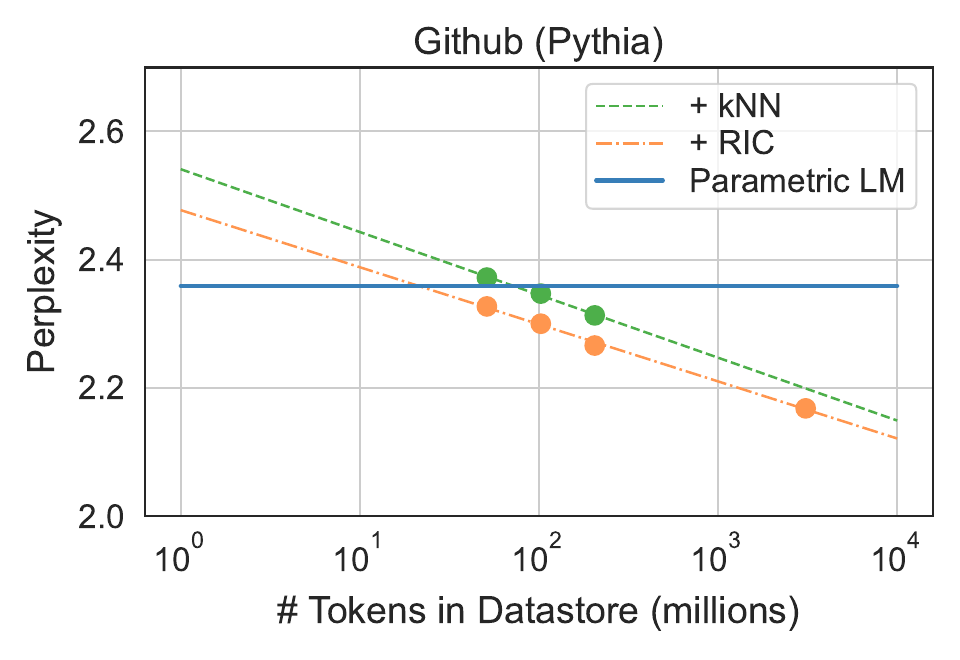}
\includegraphics[width=0.42\columnwidth]{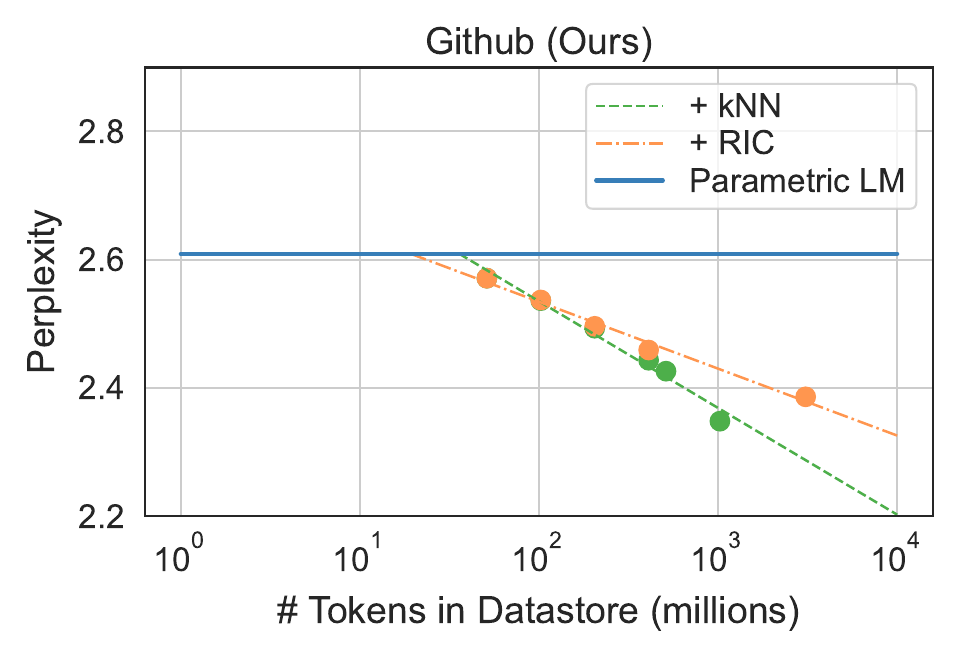}
\caption{
    Comparison between parametric LM, \riclm\, and \knnlm\ on five domains, with Pythia (left) and \modelname\ \PD\SW\ (right), respectively.
    Perplexity on random 128K tokens from the validation data reported.
}
\label{fig:scale-datastore-all}
\end{figure*}

Figure~\ref{fig:scale-datastore-all} reports the results: Pythia on the left, \modelname\ (\PD\SW) on the right.
Results show that both Pythia and \modelname\ see consistent improvements from \knnlm\ and \riclm\ as the datastore gets larger, although the slope is larger with \modelname\ than with Pythia. Again consistent to findings in \S\ref{subsec:results-nonparametric}, \knnlm\ scales better than \riclm\ does, resulting in \knnlm\ outperforming \riclm\ with a reasonably large datastore in most cases (with an exception of Pythia on Github, where \riclm\ outperforms \knnlm\ with a reasonable size of a datastore).

\begin{wrapfigure}{r}{0.36\textwidth}
  \begin{center} \myfontsize
  \setlength{\tabcolsep}{4pt}
    \begin{tabular}{lrl}
        \toprule
            Method & PPL & Disk use \\
        \midrule
            Param-only & 19.7 & 0.0 \\
        \midrule
            No approximation & 16.4 & 1.0 \\
            Quantized (4x) & 16.6 & 0.25 \\
            Quantized (8x) & 16.6 & 0.125 \\
            Quantized (16x) & 16.8 & 0.0625 \\
            IVFPQ approximation & 16.8 & 0.0178 \\
        \bottomrule
    \end{tabular}
  \end{center}
  \vspace{-.4em}
  \captionof{table}{
    Ablations on approximation methods on the validation data of Wikipedia, using the LM trained on \PD\SW\ and the datastore consisting of 51.2 million tokens (5\% of the datastore in the main experiments).
    Relative disk memory usage reported (considering {\em no approximation} as 1.0).
  } \label{tab:ablation-approximation}
\end{wrapfigure}

\tightparagraph{Effect of different approximation methods for L2 distance.}
Prior work~\citep{khandelwal2020generalization} typically uses approximate nearest neighbor search to find the top $k$ nearest neighbors, and then computes the exact L2 distance using the original vectors.
However, this may be inefficient in disk memory usage and run-time speed, due to needing to store large, original vectors and access them on-disk.
We thus explore a few alternatives: (1) quantizing the original vectors to compute the L2 distance (but less aggressively than quantization for the nearest neighbor search index, thus it provides different levels of approximations for search and for L2 distance), or (2) completely dropping the original vectors and relying on approximated L2 distance from the FAISS index with aggressive quantization.
Based on Table~\ref{tab:ablation-approximation}, all approximation methods only marginally affect performance.
For the rest of our experiments, we use the most aggressive approximation that completely drops the original embeddings at the cost of about $0.5\%$ lose in performance while using $<2\%$ of the memory footprint. Future work may study more accurate and efficient approximation methods.

\begin{wrapfigure}{r}{0.4\textwidth}
  \begin{center} \myfontsize
    \setlength{\tabcolsep}{4pt}
    \vspace{-2.5em}
    \begin{tabular}{lrr}
        \toprule
            Method & PPL & \# tokens/s \\
        \midrule
            Param-only & 19.7 & 1828.6 \\
        \midrule
            \riclm\ (51.2M) & 19.3 & 812.7\\
            \riclm\ (102.4M) & 19.2 & 731.4\\
            \riclm\ (204.8M) & 19.1 & 588.5 \\
            \riclm\ (409.6M) & 18.9 & 478.5 \\
            \riclm\ (1,178M) & 18.9 & 419.7\\
        \midrule        
            \knnlm\ (51.2M) & 16.8 & 184.2 \\
            \knnlm\ (102.4M) & 16.3 & 112.0 \\
            \knnlm\ (204.8M) & 15.7 & 59.3 \\
            \knnlm\ (409.6M) & 15.0 & 31.8 \\
            \knnlm\ (1,024M) & 14.2 & 14.2 \\
        \midrule
            \knnlm\ (102M, $p=1$) & 16.7 & 560.8 \\
            \knnlm\ (1,024M, $p=1$) & 14.6 & 71.1 \\
            \knnlm\ (1,024M, $p=2$) & 14.4 & 45.5 \\
            \knnlm\ (1,024M, $p=4$) & 14.2 & 27.0 \\
        \bottomrule
    \end{tabular}
  \end{center}
  \vspace{-.4em}
  \captionof{table}{Comparison in runtime speed (\# tokens per second) on the validation data of Wikipedia. $p$ indicates the number of probe, one of the hyperparameters in fast nearest neighbor search ($p=8$ in all experiments in the paper if not specified otherwise).
  } \vspace{-2.5em} \label{tab:runtime}
\end{wrapfigure}

\tightparagraph{Runtime speed.}
Table~\ref{tab:runtime} presents the runtime speed of the parametric LM, \riclm, and \knnlm\ on the Wikipedia validation set. Speed is reported in tokens per second with a batch size of 1 using a single NVIDIA RTX 6000 GPU.

The results show that the parametric LM is notably faster than both \riclm\ and \knnlm, and \riclm\ is faster than \knnlm. Speed is slower as the datastore gets larger (for both \riclm\ and \knnlm) and the nearest neighbor search gets less accurate (for \knnlm; indicated by the number of probe $p$). \knnlm\ can eventually match \riclm's speed while surpassing its performance by using a smaller datastore and less accurate search, i.e., when using 102M tokens with $p=1$.

We note that the machine used for benchmarking speed has a very slow IO speed, leading to an underestimation of both \riclm\ and \knnlm's runtime speed, and the comparison can significantly vary based on the hardware.
However, it is still important to note that \knnlm\ is substantially slower than a parametric LM, leaving room for potential future improvements.

\begin{table}[t]
    \myfontsize \centering
    \begin{tabular}{p{13.5cm}} 
        \toprule
            \testPrefix\ \\
            include ‘../lib/admin.defines.php’; \\
            include ‘../lib/admin.module.access.php’; \\
            include ‘../lib/admin.smarty.php’; \\
            if (! has\_rights ( \\
            \testContinuation\
            \cn{AC}X\_BILLING)) \{ Header … \\
            \retrievedPrefix\  \\
            (...) \\
            * You should have received a copy of the GNU Affero General Public License \\
            * along with this program. If not, see $<$http://www.gnu.org/licenses/$>$. \\
            * \\
            * \\
            **/ \\
            if  (! has\_rights ( \\
            \retrievedContinuation\ \cn{AC}X\_ACCESS)) \{ Header ... \\
        \midrule
            \testPrefix\ \\
            0x5f \#define S5K4AA\_DEFAULT\_BRIGHTNESS 0x10 \\
            /******************/ \\
            /* Kernel \\
            \testContinuation\ \cn{module} parameters */ extern int force\_sensor; ... \\
            \retrievedPrefix \\
            * Copyright © 2011-2013 Jozsef Kadlecsik $<$kadlec@blackhold.kfki.hu$>$ \\
            * \\
            * This program is free software; you can redistribute it and/or modify \\
            * it under the terms of the GNU General Public License version 2 as  \\
            * published by the Free Software Foundation. \\
            */ \\
            /* Kernel \\
            \retrievedContinuation\ \cn{module} implementing an IP set type: … \\
        \midrule
            \testPrefix\ … Mark or credit about hedge funds? Sara \\
            Sara Shackleton \\
            Enron North America Corp. \\
            \texttt{[Address]} \\
            \texttt{[Phone number]} \\
            \texttt{[Email address]} \\
            --- Forwarded by Sara Shacleton/HOU/ECT on 01/2023/2022 05:41PM --- \\
            Tana \\
            \testContinuation\ \cn{Jones}~~~~~12/14/2000 \\
            \retrievedPrefix\
            ... Food will be provided! Tana: Please feel free to extend the invitation to any Enron employees who may be interested in te presentation. 1st come, 1st serve. Thanks, Sylvia. --- Forwarded by Sylvia Hu/Corp/Enron on 07/14/2000 03:17PM --- Tana \\
            \retrievedContinuation\ \cn{Jones}@ECT. 07/13/2000
            \\
        \midrule
            \testPrefix\ Ken Lay and Jeff Skilling were interviewed on CNNfn to discuss the succession of Jeff to CEO of Enron. (...) and then choose “Enron’s Succession Plan.”. The interview will be available every 15 minutes \\
            \testContinuation\ \cn{through} Friday, Dec. 15. \\
            \retrievedPrefix\ Did you miss Jeff on CNBC “Street Signs” yesterday? Not to worry. (...) and then choose $>$ “Skilling CNBC.”. The interview will be available every ten minutes \\
            \retrievedContinuation\ \cn{through} $>$ Wednesday, Dec. 6. \\
        \midrule
            \testPrefix\ … The teams toured the city, explored west Edmonton mall and also got to take in an Oilers practice where they met German hockey star Leon \\
            \testContinuation\ \cn{Draisait}l \\
            \retrievedPrefix\ One minute and 19 seconds later, Cannor McDavid took a pass from Leon \\
            \retrievedContinuation\ \cn{Draisait}l  \\
        \midrule
            \testPrefix\ ... Foley on RAW’s run-time issues. Claiming that having the show run so late is one of the reasons why the final hour of RAW tends to struggle, Foley didn’t end there. ``No one else at 10:30pm is a \\
            \testContinuation\ \cn{PG} show. I won't say that across \\
            \retrievedPrefix\ … way to the ring' podcast Foley cited RAW's duration and RG rating as hindrances to the show’s popularity. Here's what he had to say: ``Sometiems we try to look into the reasons why the third hour doesn’t perform as well as the first two, and I'm like 'well that’s because people go to bed! No one else at 10:30pm is a \\
            \retrievedContinuation\ \cn{PG} show. I won't say that across \\
        \bottomrule
    \end{tabular}
    \caption{Qualitative examples of retrieved context of our model.
        \cn{Red underline text} indicates the next token that immediately follows the prefix.
        The first two are from Github; the next two are from Enron Emails; and the last two are from CC News.
    }\label{tab:qualitative-examples}
\end{table}

\tightparagraph{Qualitative examples.}
Figure~\ref{tab:qualitative-examples} provides six qualitative examples on the top-1 context retrieved by \modelname-based \knnlm.
The model is able to assign a high probability to the ground truth token by retrieving highly relevant context, e.g., given the context (hockey) and the first name of the player, being able to retrieve the last name of the player, given the context (a show and its host), being able to complete the quote.
These examples also highlight that a nonparametric approach addresses specific legal risks that we have discussed earlier, e.g., it assigns per-token attribution for free, and can provide a copyright notice when part of copyrighted text is being used for the probability distribution.

\begin{table}  
\myfontsize 
\setlength{\tabcolsep}{4pt}
\centering
\begin{tabular}{lrrrrrr}
\toprule
Dataset & BookCorpus2 & Books3 & Enron Emails & FreeLaw & Github & Gutenberg (PG-19) \\
\midrule
ccby\_law & 0.02 & 0.04 & 0.02 & 0.08 & 0.02 & 0.03 \\
ccby\_s2orc & 0.05 & 0.10 & 0.03 & 0.06 & 0.05 & 0.07 \\
ccby\_stackexchange & 0.05 & 0.07 & 0.03 & 0.04 & 0.16 & 0.05\\
ccby\_stackoverflow & 0.03 & 0.05 & 0.01 & 0.03 & 0.07 & 0.03 \\
ccby\_wikinews & 0.07 & 0.15 & 0.02 & 0.08 & 0.03 & 0.09 \\
ccby\_wikipedia & 0.05 & 0.11 & 0.02 & 0.06 & 0.03 & 0.08  \\
pd\_books & 0.13 & 0.26 & 0.03 & 0.07 & 0.04 & 0.33  \\
pd\_law & 0.05 & 0.10 & 0.02  & 0.35 & 0.03 & 0.07 \\
pd\_news & 0.06 & 0.14 & 0.02 & 0.07 & 0.02 & 0.08 \\
pd\_s2orc & 0.08 & 0.15 & 0.04 & 0.08 & 0.05 & 0.13  \\
sw\_amps\_math & 0.01 & 0.02 & 0.01 & 0.01 & 0.02 & 0.01  \\
sw\_dm\_math & 0.00 & 0.01 & 0.00 & 0.01 & 0.01 & 0.01 \\
sw\_github & 0.04 & 0.04 & 0.03 & 0.03 & 0.24 & 0.04\\
sw\_hackernews & 0.09 & 0.18 & 0.03 & 0.06 & 0.06 & 0.10\\
sw\_ubuntu\_irc & 0.12 & 0.11 & 0.06 & 0.04 & 0.08 & 0.09 \\
\midrule
\small
Dataset & OpenWebText2 & PhilPapers & Wikipedia (en) & cc-news & new-amazon & HackerNews  \\
\midrule
ccby\_law & 0.02 & 0.03 & 0.05 & 0.05 & 0.02 & 0.03 \\\
ccby\_s2orc & 0.05 & 0.11 & 0.02 & 0.06 & 0.06 & 0.07 \\
ccby\_stackexchange & 0.06 & 0.08 & 0.04 & 0.05 & 0.06  & 0.10 \\
ccby\_stackoverflow & 0.05 & 0.07 & 0.02 & 0.03 & 0.07 & 0.06 \\
ccby\_wikinews & 0.09 & 0.22 & 0.02 & 0.04 & 0.09 & 0.08 \\
ccby\_wikipedia & 0.06 & 0.17 & 0.02 & 0.07 & 0.07 & 0.06\\
pd\_books & 0.16 & 0.15 & 0.03 & 0.11 & 0.11 & 0.09\\
pd\_law & 0.06 & 0.10 & 0.02 & 0.06 & 0.06  & 0.05\\
pd\_news & 0.08 & 0.21 & 0.02 & 0.06 & 0.09 & 0.06 \\
pd\_s2orc & 0.08 & 0.13 & 0.03 & 0.09 & 0.08 & 0.09\\
sw\_amps\_math & 0.01 & 0.02 & 0.01  & 0.01 & 0.01 & 0.02\\
sw\_dm\_math & 0.00 & 0.01 & 0.00 & 0.00 & 0.00  & 0.00 \\
sw\_github & 0.04 & 0.04 & 0.03 & 0.05 & 0.03  & 0.06 \\
sw\_hackernews & 0.14 & 0.20 & 0.04 & 0.06 & 0.16  & 0.19 \\
sw\_ubuntu\_irc & 0.13 & 0.10& 0.10& 0.10 & 0.08  & 0.18\\
\bottomrule
\end{tabular}
\caption{Unigram and bigram overlap between the domain of the Pile validation data and the domains of \dataname.}\label{tab:ngram-overlap-ours-pile}
\end{table}

\end{document}